\documentclass{style/hcrg-nasa}

\issuer{Harvard Computational Robotics Group}
\edition{Preprint}
\logofile{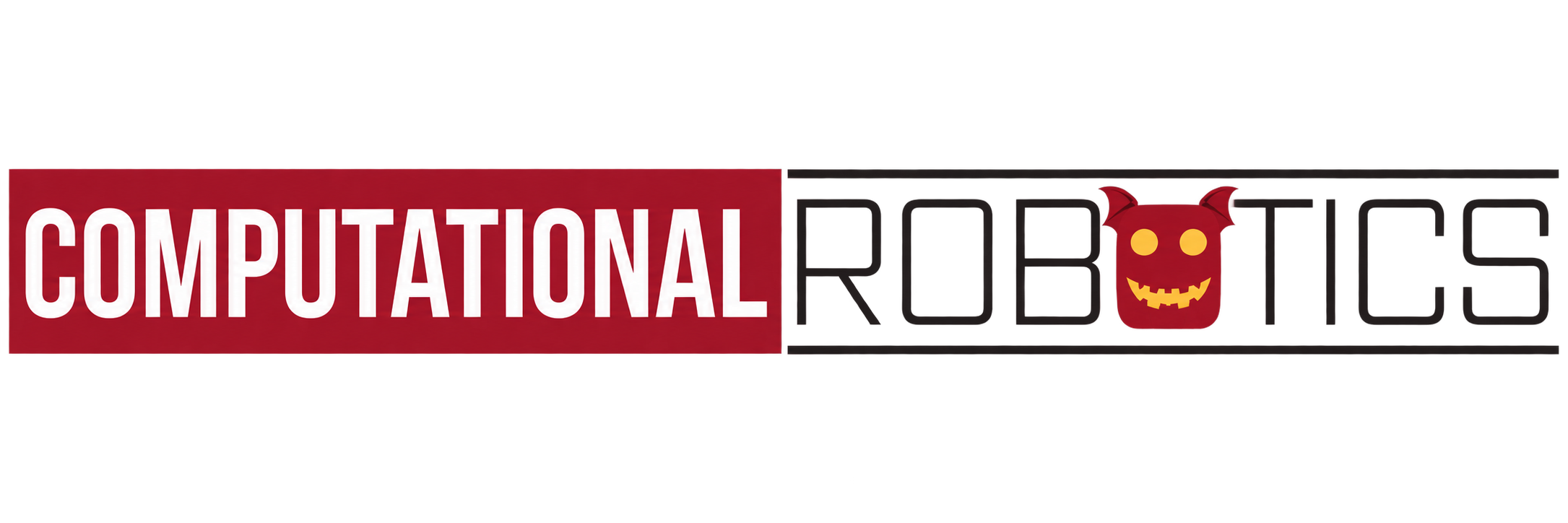}
\logotrim{0bp 220bp 0bp 224bp}

\usepackage[utf8]{inputenc}
\usepackage{algorithm}

\makeatletter
\renewcommand{\For}{\csname ALG@cmd@\ALG@L @For\endcsname}
\renewcommand{\While}{\csname ALG@cmd@\ALG@L @While\endcsname}
\renewcommand{\If}{\csname ALG@cmd@\ALG@L @If\endcsname}
\renewcommand{\Else}{\csname ALG@cmd@\ALG@L @Else\endcsname}
\renewcommand{\Return}{\csname ALG@cmd@\ALG@L @Return\endcsname}
\makeatother

\newcommand{\grasp}{GRASP}

\newtheorem{proposition}{Proposition}
\newtcolorbox{propositionbox}{
  enhanced, sharp corners, frame hidden,
  colback=crimson!5!white, borderline west={2pt}{0pt}{crimson},
  boxrule=0pt, boxsep=0pt,
  left=10pt, right=10pt, top=5pt, bottom=5pt,
  before skip=8pt, after skip=10pt
}

\newcommand{\OverallEvaluated}{1352}

\newcommand{\OverallSuccessRate}{94.2}
\newcommand{\OverallPosMean}{2.87}
\newcommand{\OverallPosMedian}{2.17}
\newcommand{\OverallRotMean}{0.117}
\newcommand{\OverallRotMedian}{0.090}
\newcommand{\MainSuccess}{1254}
\newcommand{\MainSuccessRate}{92.8}
\newcommand{\MainPosMean}{3.19}
\newcommand{\MainPosMedian}{2.31}
\newcommand{\MainRotMean}{0.127}
\newcommand{\MainRotMedian}{0.098}

\newcommand{\FullDurationMean}{8.16}
\newcommand{\FullDurationMedian}{5.80}

\newcommand{\NoDevEvaluated}{1345}
\newcommand{\NoDevSuccess}{1268}
\newcommand{\NoDevSuccessRate}{94.3}
\newcommand{\HotAllAUC}{81.54}
\newcommand{\HotAllTasks}{5,400}
\newcommand{\HotBimanualAUC}{75.83}
\newcommand{\HotRightAUC}{84.87}
\newcommand{\HotLeftAUC}{85.44}

\title{GRASP: Gradient-Guided Sampling for Contact-Rich Physics-Based Motion Retargeting}
\author[1*]{Yang Liu}
\author[1*]{Tong Wang}
\author[1]{Yiming Bao}
\author[1]{Heng Yang}
\affiliation[1]{School of Engineering and Applied Sciences, Harvard University}
\contribution[*]{Equal contribution}

\hypersetup{
  pdftitle={GRASP: Gradient-Guided Sampling for Contact-Rich Physics-Based Motion Retargeting},
  pdfauthor={Yang Liu, Tong Wang, Yiming Bao, Heng Yang},
  pdfkeywords={Physics-based motion retargeting, trajectory optimization,
    sequential Monte Carlo, model predictive control, robot learning}
}

\abstract{
We study physics-based retargeting of human demonstrations for robot
learning: finding robot controls that reproduce reference motions under
contact dynamics. We propose \grasp{} (Gradient Recovery from Archived
Samples for Proposals), which combines sampling-based exploration with
gradient-guided refinement without requiring a differentiable simulator.
GRASP-TO extends tempered sequential Monte Carlo by fitting local
quadratic cost models to archived rollouts and using their gradients to
guide Markov chain Monte Carlo rejuvenation. Metropolis--Hastings
correction preserves the target distribution for each fixed proposal
kernel, and a population analysis connects the recovered gradients to
Gaussian smoothing. GRASP-MPC scales this optimizer to long demonstrations
through warm-started receding-horizon planning. Experiments on dexterous
hands and humanoids show accurate retargeting with a shared tracking
objective and no task-specific contact rewards, including
\MainSuccessRate{}\% success on \OverallEvaluated{} bimanual OakInk2
demonstrations. The refined references improve PPO tracking in pen
spinning and humanoid--object interaction. We also demonstrate dexterous
manipulation and whole-body humanoid motion on real hardware.
}
\keywords{Physics-based motion retargeting, trajectory optimization,
  sequential Monte Carlo, model predictive control, dexterous manipulation, humanoid}
\website{https://computationalrobotics.seas.harvard.edu/GRASP}
  {\texttt{https://computationalrobotics.seas.harvard.edu/GRASP}}

\begin{document}
\maketitle

\par\vspace{4pt}
\noindent\begin{minipage}{\linewidth}
  \centering
  \makebox[\linewidth][c]{\includegraphics[width=1.06\linewidth]{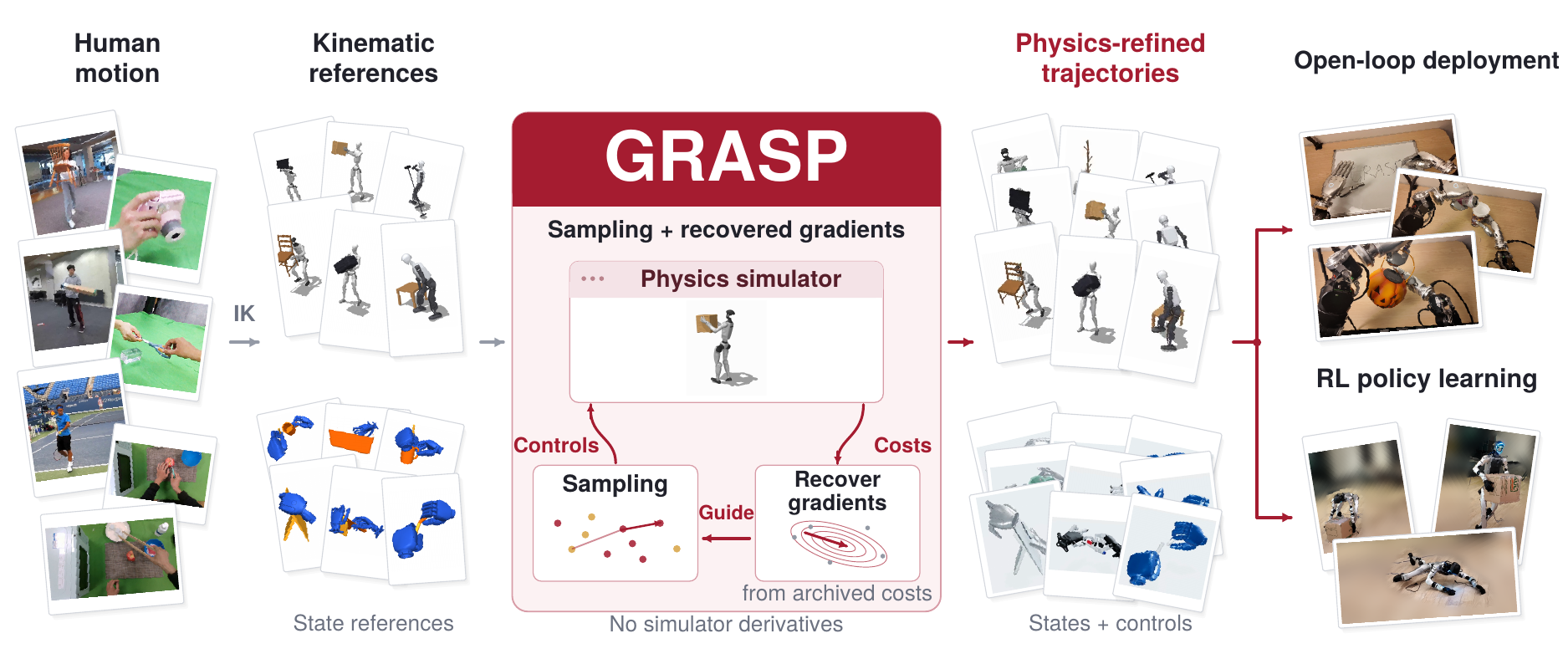}}
  \captionsetup{skip=4pt}
  \captionof{figure}{\textbf{GRASP in the human-to-robot pipeline.}
  GRASP combines sampling and recovered gradients through simulator feedback,
  refining kinematic references for open-loop deployment and RL tracking.}
  \label{fig:teaser}
\end{minipage}
\par\clearpage

\section{Introduction}
Human demonstrations provide a rich source of reference motions for
robot learning, capturing coordinated movements and interactions that
would be difficult to specify by hand. Retargeting these demonstrations
to robots with different body geometries and actuation capabilities
provides references for expressive humanoid control
\citep{cheng2024exbody,he2024h2o,he2025asap} and dexterous manipulation
\citep{qin2022dexmv}. The quality of the resulting references matters:
motions that a robot can physically realize provide a stronger basis
for learning reliable tracking policies.

Retargeting methods broadly fall into two categories.
\emph{Kinematic retargeting} maps human motion onto robot configurations,
accounting for differences in geometry and joint limits and, in some
methods, preserving spatial relationships with objects and terrain
\citep{araujo2025gmr,yang2025omniretarget}. These geometric constraints
alone do not ensure that the robot can generate the forces needed to
execute the motion. A hand may follow an object without establishing
a grasp, or a humanoid may attempt a movement that its ground contacts
cannot support. \emph{Physics-based retargeting}, the focus of this
paper, addresses this gap by optimizing robot actions in simulation
\citep{yang2025physicsgen,pan2025spider,paliwal2026do,dhedin2026dynaretarget}.
It seeks to reproduce demonstrated movements and interactions while
satisfying the simulated dynamics, making it especially important for
contact-rich manipulation and whole-body motion.

Physics-based retargeting
optimizes robot controls to track robot-state references derived from
human motion and, when present, demonstrated object motion, subject to
the dynamics of a chosen physics simulator (cf.\ Section~\ref{sec:problem}). Two common approaches to
this trajectory optimization problem are sampling-based search and
gradient-based nonlinear programming. Sampling methods such as MPPI and
cross-entropy-method variants evaluate populations of control sequences
through simulator rollouts
\citep{williams2017mppi,rubinstein1999cem,pinneri2021icem}; they support
broad exploration and have enabled large-scale physics-based retargeting
\citep{pan2025spider,paliwal2026do}. Gradient-based methods exploit local
sensitivities and can offer faster local convergence. For example,
\citet{esteban2026shooting} formulate dynamic motion retargeting as a
multiple-shooting nonlinear program using simulator transition
derivatives. These approaches offer complementary advantages, but
combining them requires useful gradient information.

Contact dynamics make this requirement difficult to meet. A small
control change can create or remove a contact and substantially alter
the subsequent motion. Many simulator interfaces expose only forward
rollouts; even when derivatives are available through regularization,
finite differences, or automatic differentiation, they can vary sharply
or fail to reveal improvements across contact modes
\citep{suh2022bundled,suh2022policy}. This motivates our central question:

\begin{tcolorbox}[
  enhanced, sharp corners, frame hidden,
  colback=crimson!5!white,
  borderline west={2pt}{0pt}{crimson},
  boxrule=0pt, boxsep=0pt,
  left=10pt, right=10pt, top=8pt, bottom=8pt,
  before skip=8pt, after skip=10pt,
  grow to left by=0pt, grow to right by=0pt
]
\emph{Can we achieve high-performance physics-based retargeting by
combining sampling-based exploration and gradient-guided refinement
without assuming differentiable simulator dynamics?}
\end{tcolorbox}

\textbf{Contributions.}
Tempered sequential Monte Carlo (TSMC) provides a principled framework
for this combination: importance reweighting and resampling progressively
favor low-cost trajectories, interleaved with gradient-guided Markov
chain Monte Carlo (MCMC) rejuvenation \citep{yang2026tempered}.
Its existing trajectory optimization formulation, however, obtains
those gradients by differentiating through the simulator and therefore
does not directly provide gradient-guided rejuvenation for
nondifferentiable dynamics. We introduce \grasp{} (\underline{G}radient \underline{R}ecovery
from \underline{A}rchived \underline{S}amples for \underline{P}roposals), which supplies this guidance by
reusing the rollout evaluations already produced during
search. Our contributions are fourfold:
\begin{itemize}
\item \textbf{GRASP-TO: trajectory optimization from recycled rollouts.}
We archive evaluated trajectory--cost pairs, including rejected
proposals, and fit local quadratic cost models in a reduced temporal
basis. Their surrogate gradients guide MCMC rejuvenation without
additional simulator queries for fitting. We establish that, for
deterministic rollout costs and fixed surrogate and proposal parameters,
the Metropolis--Hastings correction preserves the original tempered
target distribution despite surrogate fitting errors. We also show
that, under a Gaussian fitting measure, the unregularized population
quadratic recovers the gradient and Hessian of a Gaussian-smoothed cost
at the fitting center, connecting rollout regression to randomized
smoothing \citep{nesterov2017random,suh2022bundled}.

\item \textbf{GRASP-MPC: retargeting long demonstrations.}
We extend GRASP-TO to a receding-horizon formulation that replans from
the simulated states actually reached. Shifted-plan warm starts,
candidate reuse, and alternating full-search and refinement windows
reduce repeated work. Joint optimization of the initial hand
configuration and first motion window couples the starting grasp to
the ensuing interaction. This applies the hybrid optimizer to long
retargeting problems using a standard state-tracking objective without
task-specific contact rewards.

\item \textbf{Extensive simulation and policy-learning evaluation.}
We evaluate GRASP-TO on smooth and contact-switching optimization
benchmarks and GRASP-MPC on dexterous-hand and humanoid retargeting.
On \OverallEvaluated{} OakInk2 demonstrations with Sharpa hands, we
achieve \MainSuccessRate{}\% success, exceeding the baseline
success rates; on \HotAllTasks{} HOT3D trajectories, we
achieve \HotAllAUC{}\% ADD-AUC. The hand-retargeting method
supports Wuji and XHand with success rates above 92\%.
On LAFAN1, AMASS, and OMOMO, we obtain lower joint and pelvis-translation
errors than GMR and SPIDER. Controlled PPO comparisons
show that the refined references improve downstream learning and
tracking for both pen spinning and humanoid--object interaction.

\item \textbf{Real-world dexterous hands and humanoid execution.}
We demonstrate open-loop playback of \grasp{}-generated trajectories
on physical dexterous hands and closed-loop execution of learned
policies trained to track refined references on humanoids.
Demonstrations include lifting and pouring with hands, and crawling,
box carrying, and box pushing with humanoids, showing how the refined
motions support both trajectory playback and policy learning for
real-world interaction.
\end{itemize}

\textbf{Outline.}
Sections~\ref{sec:related-work} and~\ref{sec:problem} review related work
and formulate physics-based retargeting.
Sections~\ref{sec:method} and~\ref{sec:mpc} develop GRASP-TO and GRASP-MPC.
Sections~\ref{sec:exp-hands} and~\ref{sec:exp-humanoid} evaluate
dexterous-hand and humanoid retargeting, including policy learning and
real-robot execution. Section~\ref{sec:conclusion} concludes.

\section{Related Work}
\label{sec:related-work}
\textbf{Motion retargeting.}
Kinematic retargeting adapts human motion to robot geometry, providing
references for humanoid control in H2O \citep{he2024h2o} and GMR
\citep{araujo2025gmr}. OmniRetarget further preserves robot--object and
robot--terrain relationships during whole-body retargeting
\citep{yang2025omniretarget}. Physics-based methods refine these
references by optimizing robot actions under simulated dynamics.
PhysicsGen combines kinematic retargeting and trajectory optimization
across embodiments \citep{yang2025physicsgen}, while DynaRetarget uses
sampling-based trajectory optimization for dynamically feasible motion
references \citep{dhedin2026dynaretarget}. Several approaches assist
contact discovery with specialized guidance: SPIDER uses virtual
contact constraints to guide retargeting for hands and humanoids
\citep{pan2025spider}, and Do-as-I-Do augments sampling-based dexterous
retargeting with rewards for pick-and-place transitions
\citep{paliwal2026do}. Shooting for Contact \citep{esteban2026shooting}
instead uses a multiple-shooting nonlinear program with simulator
derivatives to retarget humanoid motion without prescribing contact
schedules. \grasp{} uses a common robot-and-object state-tracking
objective without virtual contact guidance or task-specific contact
rewards. The optimizer discovers the contacts needed to reproduce the
demonstrated interaction through forward simulation, without requiring
dynamics derivatives.

\textbf{Optimization through contact.}
Three broad approaches address the nonconvexity and nonsmoothness of
contact-rich trajectory optimization. \emph{Local gradient-based
methods} include contact-invariant optimization \citep{mordatch2012hand}
and the direct method of \citet{posa2014direct}, which jointly optimizes
motion and contact forces through a complementarity-constrained
nonlinear program solved by sequential quadratic programming.
More recent methods include CRISP's sequential convex optimization
\citep{li2025crisp}, IMPACT's augmented-Lagrangian formulation
\citep{li2026impact}, and the Contact Trust Region \citep{suh2026ctr}.
These methods can provide efficient local refinement, but require
derivative information and can remain sensitive to initialization and
local minima. \emph{Sampling-based methods}, such as MPPI
\citep{williams2017mppi} and iCEM \citep{pinneri2021icem}, explore
candidate trajectories without differentiating the dynamics. However,
local convergence can be slow and require many rollouts in
high-dimensional control spaces. \emph{Global optimization through
convex relaxations} provides lower bounds on the optimal cost and can
produce feasible plans with certified suboptimality gaps.
\citet{graesdal2024tight} combine semidefinite relaxations with graphs
of convex sets for contact-mode selection in quasi-static planar
pushing. \citet{kang2025global} exploit polynomial and robotics-specific
sparsity to construct higher-order semidefinite relaxations for
contact-rich planning. For polynomial trajectory optimization, STROM
combines sparse semidefinite relaxations with a GPU-accelerated solver
\citep{kang2026fast,groudiev2025cuadmm}. Despite substantial gains from sparsity and
parallelization,
large relaxations remain computationally expensive as the problem
dimension and number of contact modes grow.
\grasp{} combines sampling-based exploration with surrogate-gradient
refinement. Its optimizer requires only batched rollout costs, allowing
it to operate with any simulator that provides this interface and to
address long, high-dimensional demonstrations through receding-horizon
optimization.

\textbf{Sampling-based optimization and gradient recovery.}
Stochastic optimizers use rollout populations in different ways:
STOMP updates trajectories using noisy rollout costs
\citep{kalakrishnan2011stomp}, and CBO studies global trajectory
optimization through interacting particles \citep{sun2026cbo}.
Sequential Monte Carlo evolves a population through intermediate target
distributions \citep{delmoral2006smc}; TSMC adapts this framework to
trajectory and policy optimization by combining tempering with
rejuvenation guided by simulator gradients \citep{yang2026tempered}.
Other hybrids combine CEM with gradient descent in MPC
\citep{bharadhwaj2020mpc} or guide MPPI with stochastic surrogate
gradients, as in SVG-MPPI \citep{honda2023svgmppi}.
When analytic derivatives are unavailable, local surrogate models
provide search directions in derivative-free optimization
\citep{conn2009dfo}, while randomized smoothing connects function
evaluations to gradients of smoothed objectives
\citep{nesterov2017random,suh2022bundled,suh2022policy}.
NFG estimates natural functional gradients of a Gaussian-smoothed
trajectory objective from cost evaluations \citep{park2026nfg}, and
WASP reuses information across successive derivative estimates
\citep{rakita2025wasp}. \grasp{} integrates rollout reuse into tempered
population search: local quadratic fits to archived costs, including
rejected proposals, guide MCMC rejuvenation, while Metropolis--Hastings
acceptance uses the true simulator costs. This couples exploration and
local refinement without additional simulator queries for fitting or
access to dynamics derivatives.

\textbf{Sampling-based inference in perception, localization, and statistics.}
Particle methods have a long history in robotics and vision
\citep{dellaert2003sample}. Monte Carlo localization represents a robot's
pose posterior with weighted samples \citep{dellaert1999monte}, while
MCMC-based particle filtering uses Markov chain updates to sample the
joint posterior of interacting visual targets \citep{khan2004mcmc}.
Annealed particle filtering applies a sequence of increasingly
concentrated fitness functions to articulated body tracking
\citep{deutscher2000articulated}. In statistics, resample--move methods
combine importance sampling and resampling with target-invariant MCMC
moves to restore particle diversity
\citep{gilks2001moving,berzuini2001resample}; SMC samplers provide a
general framework for sampling through sequences of intermediate
distributions, including annealed targets \citep{delmoral2006smc}.
TSMC and \grasp{} build on these population-based inference mechanisms:
particles represent candidate control trajectories, and the target
sequence is indexed by an artificial inverse temperature that
progressively favors lower rollout costs. Thus, their use of
resampling and MCMC rejuvenation connects directly to established
sampling methods beyond trajectory optimization.

\textbf{Data-driven proposals and surrogate-assisted MCMC.}
Informative proposals are also well established in this literature.
\citet{oh2005datadriven} learn temporal cue models to guide
proposals over discrete mode sequences in switching linear dynamical
systems, while \citet{ranganathan2005datadriven} use geometric information
to guide split--merge proposals over topological maps. These methods
construct data-dependent discrete moves without estimating gradients
of the target density. For continuous targets,
\citet{rasmussen2003gaussian} uses a Gaussian-process surrogate to guide
Hamiltonian Monte Carlo trajectories, with Metropolis acceptance based
on the original target. Local polynomial and Gaussian-process
approximations have also been used to accelerate asymptotically exact
MCMC \citep{conrad2016accelerating}. \grasp{} adapts surrogate-guided
proposal construction to contact-rich trajectory optimization by
fitting local quadratic models online to archived rollout costs in
reduced temporal coordinates. The recovered gradients guide MCMC
rejuvenation within TSMC, while true simulator costs enter the
Metropolis--Hastings correction for each fixed proposal kernel.

\section{Problem Formulation: Physics-Based Motion Retargeting as Dynamics-Constrained Trajectory Tracking}
\label{sec:problem}
Physics-based motion retargeting seeks robot controls that reproduce a
human demonstration under simulated dynamics. A kinematically retargeted
demonstration provides reference states
$\bar x_{0:T}=(\bar x_0,\ldots,\bar x_T)$ over $T$ simulation steps.
At step $t$, the state is $x_t=(q_t,v_t)\in\mathcal X$, where
$q_t\in\mathcal Q$ is the combined robot and object configuration,
$v_t\in\mathbb R^{n_v}$ contains their generalized velocities, and
$\mathcal X=\mathcal Q\times\mathbb R^{n_v}$. The configuration space
$\mathcal Q$ includes robot joint positions and robot and object poses.
The robot command $u_t\in\mathbb R^m$ has $m$ control channels;
in our retargeting experiments, these commands are position targets.
The simulator transition
$f:\mathcal X\times\mathbb R^m\to\mathcal X$ includes the actuation and
contact dynamics.

Given an initial state $x_{\rm init}$, we optimize the command sequence
$u_{0:T-1}=(u_0,\ldots,u_{T-1})$ to track the demonstration:
\begin{equation}
 \begin{aligned}
 \min_{u_{0:T-1}}\; J(u_{0:T-1})
 &=\sum_{t=1}^{T}\ell(x_t;\bar x_t)
   +w_{\rm term}\ell(x_T;\bar x_T),\\
 \text{s.t.}\quad x_{t+1}&=f(x_t,u_t),\quad t=0,\ldots,T-1,\\
 x_0&=x_{\rm init}.
 \end{aligned}
 \label{eq:cost}
\end{equation}
Here $\ell:\mathcal X\times\mathcal X\to\mathbb R_{\geq0}$ is the
stage tracking loss, and $w_{\rm term}\geq0$ gives additional weight to
tracking at the horizon endpoint. We use the same loss for the stage
and terminal contributions:
\begin{equation}
 \ell(x_t;\bar x_t)
 =\|W_q(q_t\ominus\bar q_t)\|_2
       +w_v\|v_t-\bar v_t\|_2,
 \label{eq:tracking}
\end{equation}
where $\bar x_t=(\bar q_t,\bar v_t)$. The operator $\ominus$ forms a
configuration-error vector in $\mathbb R^{d_q}$, using relative rotations
for orientation components. The diagonal matrix
$W_q\in\mathbb R^{d_q\times d_q}$ assigns nonnegative weights to object
pose, robot base pose, and joint errors, while the scalar $w_v\geq0$
weights velocity errors. The kinematic reference need not satisfy the
simulator dynamics. Objects are unactuated and move under simulated
contact and gravity, so reproducing their demonstrated motion requires
the optimizer to discover the necessary contacts. Changes in the
controls can alter these contacts, making the rollout cost nonsmooth
even with a simple tracking loss.

\begin{figure}[htbp]
 \centering
 \includegraphics[width=\linewidth]{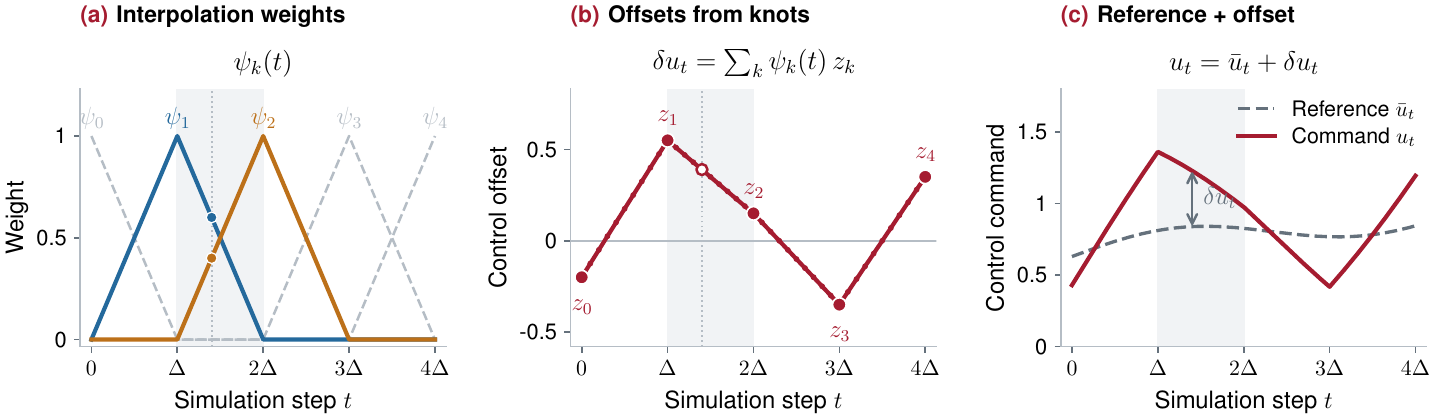}
 \caption{Control-knot parameterization for one control channel, with
 $K=5$ and illustrative values. (a) Triangular weights interpolate between
 neighboring knots; the shaded interval highlights two overlapping weights.
 (b) The optimized offsets $z_k$ (large dots) define an offset at every
 simulation step by linear interpolation. (c) Adding this offset to the
 reference command produces $u_t$. The interpolation applies
 componentwise to all $m$ control channels. The plot ends at the last
 knot; its offset is held over the final interval.}
 \label{fig:control-knots}
\end{figure}

To reduce the number of decision variables, we parameterize the
commands by $K$ control-offset knots $z_k\in\mathbb R^m$ and stack them
into $z=(z_0^\top,\ldots,z_{K-1}^\top)^\top\in\mathbb R^{Km}$:
\begin{equation}
 u_t(z)=\bar u_t+\sum_{k=0}^{K-1}\psi_k(t)\,z_k,\qquad
 \psi_k(t)=\max\bigl(0,\,1-|\min(t/\Delta,K-1)-k|\bigr).
 \label{eq:knots}
\end{equation}
where $\bar u_t\in\mathbb R^m$ contains the reference position targets,
we use $K$ knots at $t_k=k\Delta$, $k=0,\ldots,K-1$, with
$\Delta=T/K$. We linearly interpolate between knots and hold
$z_{K-1}$ over the final interval $[T-\Delta,T)$.
Figure~\ref{fig:control-knots} illustrates this parameterization
for one control channel. We write $J(z)=J(u_{0:T-1}(z))$ for the resulting objective
on $\mathbb R^{Km}$. Our optimizer accesses this objective only through
batched simulator rollouts; it requires no derivatives of $f$.

We focus on two settings, distinguished by whether the controls are
replanned as the motion unfolds.

\textbf{Open-loop trajectory optimization (TO).}
In the open-loop case, we solve Eq.~\eqref{eq:cost} once from
$x_{\rm init}$ and execute the resulting command sequence without
replanning. Each candidate is evaluated over the full horizon, so both
the rollout length and the number of knots grow with the demonstration
duration.

\textbf{Closed-loop model predictive control (MPC).}
In the closed-loop case, we solve the same tracking problem over a
shorter window initialized at the current simulated state, with the
terminal contribution applied at the window endpoint. We execute only
a prefix of the selected command sequence and then replan from the
state reached. This receding-horizon procedure closes the feedback loop
around the simulated motion during retargeting, limits the size of each
search, and adapts subsequent commands to the contacts actually realized.

Section~\ref{sec:method} develops GRASP-TO for the open-loop problem.
Section~\ref{sec:mpc} then uses this solver within GRASP-MPC to retarget
long demonstrations through closed-loop replanning.

\section{GRASP for Trajectory Optimization}
\label{sec:method}
We first develop the open-loop solver, which searches over the knot
offsets $z\in\mathbb R^{Km}$. We use a \emph{single-shooting}
formulation: each candidate defines the command sequence
$u(z):=u_{0:T-1}(z)$ through Eq.~\eqref{eq:knots}, and the simulator
unrolls $x_0(z)=x_{\rm init}$ and
$x_{t+1}(z)=f(x_t(z),u_t(z))$ over the full horizon.
Substituting these states into Eq.~\eqref{eq:cost} makes the entire cost
a function of $z$ alone; we use the shorthand $J(z):=J(u(z))$.
Thus, the states are determined by the rollout, and the dynamics are
satisfied by construction. The solver accesses $J$ through batched
simulations.

\subsection{Preliminaries: Tempered Sequential Monte Carlo}
\label{sec:tsmc}
We review tempered sequential Monte Carlo (TSMC)
\citep{delmoral2006smc,yang2026tempered}, which turns trajectory
optimization into sampling from a sequence of distributions that
increasingly favor low-cost controls.

\textbf{From optimization to inference.}
We place a Gaussian autoregressive (AR(1)) prior $p_0(z)$ on the knot
offsets, independently per control channel, with coefficient
$\rho\in[0,1)$ and positive per-channel, per-knot exploration scales
(Appendix~\ref{app:prior}). This prior has full support on
$\mathbb R^{Km}$ and favors temporally smooth perturbations of the
reference controls. Following \citet{yang2026tempered}, consider the
distributional objective
\begin{equation}
 \min_p\;\mathbb E_{z\sim p}[J(z)]
       +\beta^{-1}D_{\mathrm{KL}}(p\|p_0),
 \label{eq:inference}
\end{equation}
where $p$ ranges over probability densities, $D_{\mathrm{KL}}$ is the
Kullback--Leibler divergence, and $\beta>0$ is the inverse temperature.
Its minimizer is the Boltzmann-tilted density $\pi_1$ in the path
\begin{equation}
 \pi_\lambda(z)=\frac{1}{Z_\lambda}
      p_0(z)\exp[-\lambda\beta J(z)],
 \qquad 0\leq\lambda\leq1.
 \label{eq:target}
\end{equation}
Here $Z_\lambda$ is the normalizing constant: $\pi_0=p_0$, and
$\pi_1$ is the desired inference target. As $\beta\to\infty$, $\pi_1$
concentrates on arbitrarily near-optimal costs, provided every sublevel
set $\{z:J(z)\leq J_*+\delta\}$ has positive prior probability for
$\delta>0$, where $J_*:=\inf_z J(z)$. Tempering advances $\lambda$
from zero to one at a fixed $\beta$ to approach this potentially sharp,
multimodal target gradually.

\textbf{Importance weighting and resampling.}
TSMC maintains $N$ candidate knot vectors, called \emph{particles},
initialized as independent draws $z^{(i)}\sim p_0$, $i=1,\ldots,N$.
The superscript indexes particles, whereas the subscript in $z_k$
indexes knots within a particle. Given an equally weighted population
approximating $\pi_\lambda$, a proposed increase to
$\lambda'\in(\lambda,1]$ yields incremental importance weights
\begin{equation}
 \widetilde w_i=\exp[-(\lambda'-\lambda)\beta J(z^{(i)})],
 \qquad w_i=\frac{\widetilde w_i}{\sum_{j=1}^N\widetilde w_j}.
 \label{eq:smc-weights}
\end{equation}
These weights are proportional to the density ratio
$\pi_{\lambda'}(z^{(i)})/\pi_\lambda(z^{(i)})$ and favor lower-cost
particles. Their concentration is measured by the effective sample size
\begin{equation}
 \mathrm{ESS}(\lambda')=\frac{1}{\sum_{i=1}^N w_i^2}\in[1,N].
 \label{eq:ess}
\end{equation}
An ESS of $N$ means uniform weights; an ESS near one means that a single
particle dominates. We choose the largest tempering increment that
keeps ESS at least $\gamma N$, for a prescribed fraction
$\gamma\in(0,1)$. Resampling then draws $N$ particles with replacement
according to $w_i$ and resets their weights to $1/N$, allocating more
computation to promising trajectories.

\textbf{MCMC rejuvenation.}
Resampling can create repeated particles. Markov chain Monte Carlo
(MCMC) moves diversify these copies and explore nearby controls while
leaving the new target $\pi_{\lambda'}$ invariant. Repeating weighting,
resampling, and rejuvenation advances the population to $\lambda=1$.
The differentiable-dynamics method of \citet{yang2026tempered} uses
Hamiltonian Monte Carlo (HMC) with gradients through simulator rollouts.
Here we use the simpler Metropolis-adjusted Langevin algorithm (MALA)
\citep{roberts1996langevin} with a Metropolis--Hastings correction
\citep{hastings1970monte}. At a fixed tempering level $\lambda$, its
gradient-based proposal is
\begin{equation}
 \begin{aligned}
 z'&=z+\tfrac{\epsilon^2}{2}M g(z)+\epsilon M^{1/2}\xi,
       \qquad \xi\sim\mathcal N(0,I_{Km}),\\
 g(z)&=\nabla\log\pi_\lambda(z)
      =\nabla\log p_0(z)-\lambda\beta\,\nabla J(z),
 \end{aligned}
 \label{eq:mala}
\end{equation}
where $\epsilon>0$ is the step size and
$M\in\mathbb R^{Km\times Km}$ is a positive-definite, prior-based
preconditioner. Writing $Q_\lambda(z'\mid z)$ for this Gaussian proposal density,
we accept the proposed particle with probability
\begin{equation}
 \alpha(z,z')=\min\!\left\{1,
 \frac{p_0(z')e^{-\lambda\beta J(z')}Q_\lambda(z\mid z')}
      {p_0(z)e^{-\lambda\beta J(z)}Q_\lambda(z'\mid z)}\right\},
 \label{eq:mh}
\end{equation}
and otherwise retain $z$.

\textbf{Asymptotic guarantees.}
For a fixed tempering schedule, target-invariant Markov kernels, and
standard resampling and integrability conditions, TSMC is consistent:
as $N\to\infty$, particle averages of bounded test functions converge
in probability to their expectations under the corresponding
$\pi_\lambda$ \citep{delmoral2006smc,yang2026tempered}.
ESS-adaptive schedules also admit consistency results under additional
regularity conditions \citep{beskos2016adaptive}.
This large-population limit controls sampling error at fixed $\beta$;
the separate limit $\beta\to\infty$ concentrates the target near optimal
costs.

The TSMC weighting and resampling steps require only cost values, and
the acceptance test combines true costs with the proposal densities.
The simulator-gradient requirement enters through the proposal drift
$\nabla J(z)$, computed by differentiating the single-shooting rollout.
Many contact simulators do not provide these derivatives, and even
when available they can be unreliable near contact transitions
\citep{suh2022bundled}. Gradient-free random-walk kernels remain valid
but can mix slowly in the high-dimensional knot space.
Section~\ref{sec:recycling} develops \grasp{} to recover useful proposal
directions from rollout values without assuming differentiable dynamics.

\subsection{Recycling Rollouts into Gradients}
\label{sec:recycling}
\begin{figure}[htbp]
 \centering
 \includegraphics[width=\linewidth]{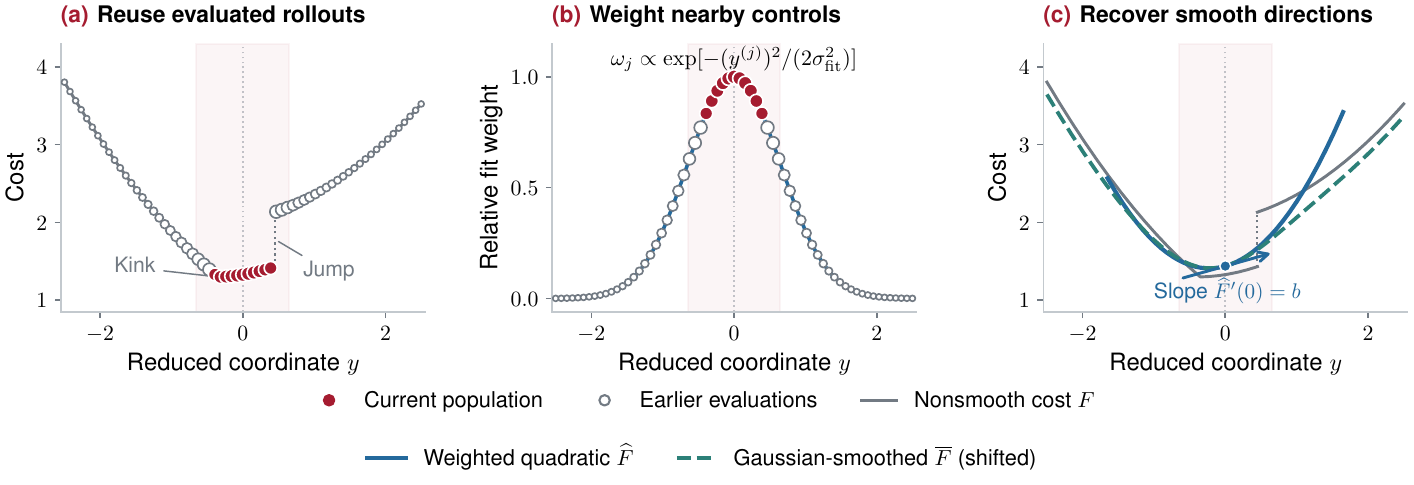}
 \caption{Recovering a smooth proposal direction from nonsmooth rollout
costs, illustrated with one reduced coordinate. (a) Current particles
(filled crimson) and earlier evaluations (open gray) both belong to the
archive; the cost has a kink and a jump. (b) Illustrative Gaussian
proximity weights with width $\sigma_{\rm fit}$ favor samples near the
current population center $y=0$; shading marks $|y|\leq\sigma_{\rm fit}$,
and marker area indicates regression weight. (c) The weighted
quadratic fit is differentiable despite these cost transitions. The
Gaussian-smoothed cost is shifted vertically to match the fitted value
at zero, allowing their local slopes and curvatures to be compared.
Proposition~\ref{prop:mollified} gives the exact derivative identities
for the ideal Gaussian population fit. All values are illustrative.}
 \label{fig:local-fit}
\end{figure}

\grasp{} replaces derivatives through the simulator with derivatives of
a local model fitted to previously evaluated controls. The same rollouts
therefore serve two purposes: they identify promising trajectories and
reveal how the cost varies around the current population.
Figure~\ref{fig:local-fit} illustrates this reuse, and
Algorithm~\ref{alg:grasp} summarizes the resulting search.

\textbf{An archive and a local coordinate system.}
Let $d:=Km$ be the dimension of a control vector. An archive
$\mathcal D=\{(z^{(j)},J(z^{(j)}))\}_{j=1}^{n_{\mathcal D}}$ stores
$n_{\mathcal D}$ evaluated controls and their costs, including proposals
rejected by the Markov kernel. Thus, $N$ denotes the current population
size, while $n_{\mathcal D}$ counts archived evaluations.
The population determines a center $\mu\in\mathbb R^d$ and a diagonal
matrix $D\in\mathbb R^{d\times d}$ with positive entries that scale each
control coordinate. We retain $1\leq r\leq K$ low-frequency discrete cosine
transform components per control channel, collected in an orthonormal
basis $B\in\mathbb R^{d\times d_r}$, where $d_r:=rm$ and
$B^\top B=I_{d_r}$. The reduced coordinates and quadratic model are
\begin{equation}
y(z)=B^\top D^{-1}(z-\mu)\in\mathbb R^{d_r},\qquad
 \widehat F(y)=a+b^\top y+\tfrac12 y^\top A y,
 \label{eq:surrogate}
\end{equation}
with intercept $a\in\mathbb R$, slope $b\in\mathbb R^{d_r}$, and
symmetric curvature matrix $A\in\mathbb R^{d_r\times d_r}$.
The low-frequency basis keeps the regression manageable while retaining
coordinated changes over time.

\textbf{A weighted quadratic fit.}
Following the local-model perspective of derivative-free optimization
\citep{conn2009dfo}, we estimate these coefficients by weighted ridge
regression:
\begin{equation}
 \min_{a,b,A=A^\top}\;
 \sum_{j=1}^{n_{\mathcal D}}\omega_j
       \bigl[J(z^{(j)})-\widehat F(y^{(j)})\bigr]^2
       +\eta\mathcal R(b,A),\qquad y^{(j)}:=y(z^{(j)}).
 \label{eq:regression}
\end{equation}
The proximity weights $\omega_j\geq0$ emphasize archived controls near
the current population in standardized coordinates. They are regression
weights, distinct from the TSMC importance weights $w_i$ in
Eq.~\eqref{eq:smc-weights}. The parameter $\eta>0$ controls a quadratic
penalty $\mathcal R$ on the slope and curvature coefficients. For finite
cost values and positive total weight, a positive-definite ridge penalty
makes the fit unique even when the samples do not determine every
coefficient. Neither the regression nor its well-posedness requires
$J$ to be continuous or differentiable.

We distinguish the reduced model $\widehat F$ from its extension to
control space, $\widehat J(z):=\widehat F(y(z))$. Its gradient follows
by the chain rule:
\begin{equation}
 \nabla_z\widehat J(z)=D^{-1}B\bigl(b+A y(z)\bigr).
 \label{eq:gradient}
\end{equation}
Fitting uses only archived rollout values and requires no additional
simulator calls. The center, scales, basis, and fitted coefficients are
held fixed throughout the associated block of Markov moves.

\textbf{Guidance with a Metropolis correction.}
We replace the cost gradient in Eq.~\eqref{eq:mala} with the recovered
direction,
\begin{equation}
 \widehat g(z)=\nabla\log p_0(z)-\lambda\beta\,\nabla_z\widehat J(z).
 \label{eq:guided}
\end{equation}
The resulting Gaussian proposal density $Q_\lambda(z'\mid z)$ must use
this same surrogate drift in both directions of the acceptance ratio
in Eq.~\eqref{eq:mh}. The simulator evaluates the true cost $J(z')$,
which enters that ratio and is appended to the archive whether the
proposal is accepted or rejected.

For any fixed surrogate, preconditioner, and step size, the corrected
kernel is reversible with respect to $\pi_\lambda$, regardless of the
surrogate's fitting error (Appendix~\ref{app:theory}). Thus, model error
can affect the efficiency of proposals without changing the invariant
distribution of that fixed kernel. 
Between move blocks, new archived
rollouts can update the model and improve subsequent guidance.

\begin{algorithm}[t]
\caption{\grasp{}: trajectory search from rollout feedback}
\label{alg:grasp}
\begin{algorithmic}[1]
\Require Cost $J$, prior $p_0$, inverse temperature $\beta$, particle count $N$, ESS fraction $\gamma$
\State Draw $z^{(i)}\sim p_0$; evaluate $J_i=J(z^{(i)})$, $i=1,\ldots,N$
\State Initialize $\mathcal D=\{(z^{(i)},J_i)\}_{i=1}^N$; set $\lambda=0$
\While{$\lambda<1$ and stage budget remains}
  \State Choose the largest $\lambda'\in(\lambda,1]$ with $\mathrm{ESS}(\lambda')\geq\gamma N$
  \State Resample particles and their costs using Eq.~\eqref{eq:smc-weights}
  \State Set $\lambda\leftarrow\lambda'$
  \State Fit $\widehat F$ from $\mathcal D$; freeze $\mu,D,B,a,b,A$, $M$, and $\epsilon$
  \For{each Markov move}
    \State Propose $z'^{(i)}$ using Eqs.~\eqref{eq:mala} and~\eqref{eq:guided}
    \State Evaluate $J(z'^{(i)})$ in parallel physics rollouts
    \State Add every $(z'^{(i)},J(z'^{(i)}))$ to $\mathcal D$
    \State Accept or reject with Eq.~\eqref{eq:mh}; update $z^{(i)}$ and $J_i$
  \EndFor
\EndWhile
\State \Return Candidate trajectories and their rollout costs
\end{algorithmic}
\end{algorithm}

\subsection{Gradient Recovery as Randomized Smoothing}
\label{sec:theory}
A quadratic fit always provides a smooth proposal field, but which
features of the nonsmooth cost does that field capture? We connect the
population regression to randomized smoothing
\citep{nesterov2017random,suh2022bundled}, separating this exact
characterization from the finite, regularized fit used in practice.

\textbf{The cost seen in reduced coordinates.}
Fix $\mu,D,B$, and let $\nu_z$ denote a population fitting measure on
control vectors: it represents both the locations of archived samples
and their proximity weights. For $Z\sim\nu_z$, write
$Y:=y(Z)$ and let $\nu$ be the distribution of $Y$. Assuming
$\mathbb E_{\nu_z}[J(Z)^2]<\infty$, define the reduced response
\begin{equation}
F(y):=\mathbb E_{\nu_z}[J(Z)\mid Y=y].
 \label{eq:reduced-response}
\end{equation}
This conditional mean accounts for variations in the coordinates that
$B$ discards. If all sampled controls lie on the slice
$Z=\mu+DBY$, it reduces to $F(y)=J(\mu+DBy)$.
Least-squares fitting to the full rollout costs is equivalent to fitting
this response: the residual variance conditional on $Y$ contributes a
constant independent of the quadratic coefficients. The unregularized
population problem is therefore
\begin{equation}
(a_*,b_*,A_*)=\arg\min_{a,b,A=A^\top}
 \mathbb E_{Y\sim\nu}\!\left[
   \bigl(F(Y)-a-b^\top Y-\tfrac12Y^\top A Y\bigr)^2\right].
 \label{eq:population-fit}
\end{equation}
A Gaussian reduced fitting measure yields an exact interpretation of
its slope and curvature coefficients.

\begin{propositionbox}
\begin{proposition}[Derivatives of a Gaussian-smoothed cost]
\label{prop:mollified}
Let $\nu=\mathcal N(0,\Sigma)$ with
$\Sigma\in\mathbb R^{d_r\times d_r}$ positive definite, and let
$F:\mathbb R^{d_r}\to\mathbb R$ be measurable with at most polynomial
growth. Define the Gaussian convolution
\begin{equation}
 \overline F(y):=\mathbb E_{Y\sim\nu}[F(y+Y)].
 \label{eq:smoothed-cost}
\end{equation}
The unique minimizer of Eq.~\eqref{eq:population-fit} satisfies
\begin{equation}
b_*=\nabla\overline F(0),\qquad
A_*=\nabla^2\overline F(0).
 \label{eq:mollified}
\end{equation}
\end{proposition}
\end{propositionbox}
The proof uses Gaussian polynomial orthogonality and differentiation
of the convolution kernel (Appendix~\ref{app:theory}). It does not
require derivatives of $F$.

The fitted gradient $b_*+A_*y$ is consequently the first-order Taylor
approximation of $\nabla\overline F(y)$ at the population center.
Gaussian convolution smooths both kinks and jumps, so neighboring cost
values can reveal a useful local direction even when the original
response has no derivative; Figure~\ref{fig:local-fit}(c) illustrates
this effect. The covariance $\Sigma$ sets the smoothing scale in
reduced coordinates; $D$ and $B$ determine how these coordinates relate
to control perturbations.

The proposition is a population identity under a Gaussian fitting
measure. A finite archive, ridge regularization, and a non-Gaussian
fitting measure can all change the recovered coefficients. It therefore
explains how value-based regression can supply smooth guidance without
identifying every empirical fit with an exact Gaussian convolution.
As the search concentrates, fitting over smaller neighborhoods can
resolve finer cost structure; the Metropolis correction continues to
use the original rollout cost at every stage.

\textbf{Optimizer validation.}
We validate GRASP-TO on acrobot swing-up and a contact-switching
pusher-slider task. To keep the main experiments focused on motion
retargeting, we defer these optimizer benchmarks and baseline comparisons
to Appendix~\ref{app:benchmarks}. We next extend the solver to long
demonstrations through receding-horizon optimization.

\section{GRASP for Model Predictive Control}
\label{sec:mpc}
\begin{figure}[htbp]
 \centering
 \includegraphics[width=\linewidth]{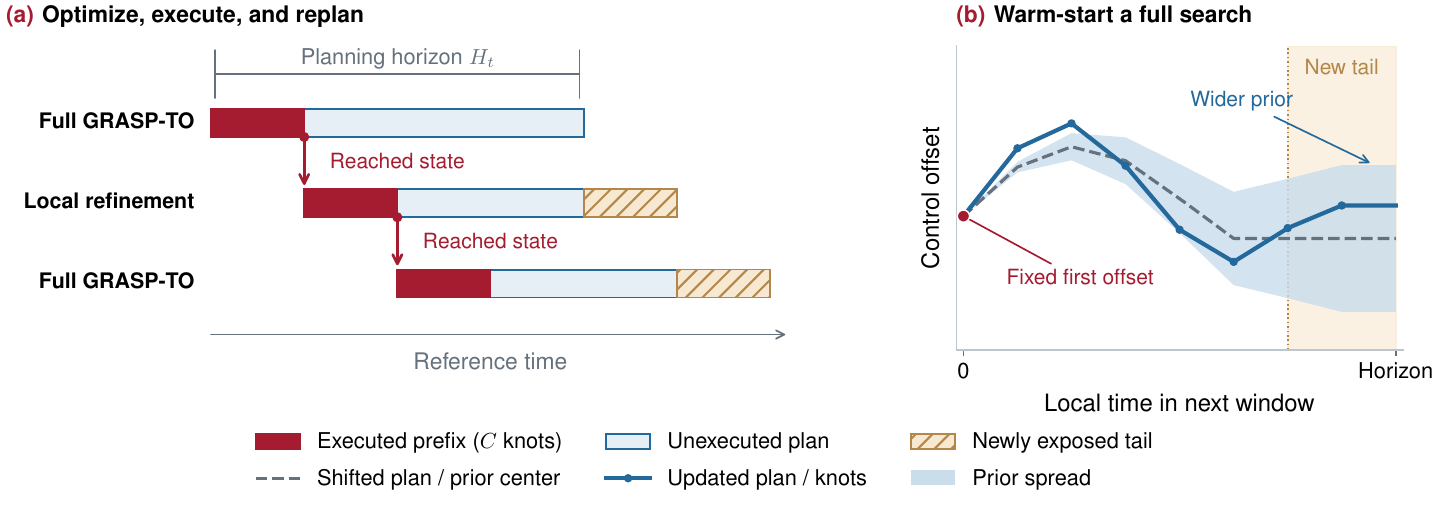}
 \caption{GRASP-MPC. (a) Each window plans over a limited horizon and
 executes only a prefix (crimson), specified by the commit count $C$.
 The next window starts from the simulated state reached by that
 prefix. Full GRASP-TO searches alternate with local refinement, reusing
 overlapping plans while exposing a new tail. (b) For a full search after
 the first window, shifted unexecuted controls center the prior, with
 the last knot repeated to fill the new tail. The shared boundary offset
 is fixed, while the prior widens toward the tail. Dots mark control
 knots; controls interpolate linearly and hold the final knot over the
 last interval. Both panels illustrate $K=8$, $C=2$; control values and
 prior widths are illustrative. Refinement windows instead reuse the
 candidate library, as detailed in Algorithm~\ref{alg:grasp-mpc}.}
 \label{fig:grasp-mpc}
\end{figure}

GRASP-TO searches over an entire control sequence from a given initial
state. For long demonstrations, both the number of decision variables
and the cost of each rollout grow with the clip duration. An early
tracking error can also change subsequent contacts and invalidate the
remainder of an open-loop plan. GRASP-MPC addresses these challenges
through the receding-horizon formulation of Section~\ref{sec:problem}:
it optimizes a bounded window, executes a short prefix, and replans from
the simulated state actually reached. We combine this feedback with
plan and particle reuse to avoid repeating a full search at every window.
Figure~\ref{fig:grasp-mpc} illustrates the overlapping windows, state
feedback, and reuse of unexecuted controls.

\subsection{Windowed Tracking and Initial Configuration}
\label{sec:mpc-window}
Let $t<T$ be the current simulation step and
$H_t\in\{1,\ldots,T-t\}$ the planning horizon in simulation steps.
Within this window, we apply the knot
parameterization of Eq.~\eqref{eq:knots} in local time, using the
corresponding segment of the reference commands and at most $K$ knots.
For a candidate offset vector $z$, single shooting generates the
predicted states $x_{t+1}(z),\ldots,x_{t+H_t}(z)$ and evaluates
\begin{equation}
 J_t(z)=\frac{1}{T}\left[
   \sum_{\tau=1}^{H_t}\ell(x_{t+\tau}(z);\bar x_{t+\tau})
   +w_{\rm term}\ell(x_{t+H_t}(z);\bar x_{t+H_t})\right].
 \label{eq:mpc-cost}
\end{equation}
Here $J_t$ denotes the current window objective, and the normalization
uses the full clip length $T$ at every replan. The terminal term
encourages tracking at the window endpoint, where errors affect the
next planning problem. Only a prefix, specified by the commit count
$C$ in knots, is executed; the remaining controls can be revised in the
next window. The horizon and executed prefix are shortened as needed
at the end of the demonstration.

\textbf{Optimizing initial contact and motion together.}
The initial hand configuration determines which contacts the first
window can establish. For dexterous-hand retargeting, we optimize it
jointly with the subsequent commands: the first control-offset knot
also parameterizes the starting hand configuration,
\begin{equation}
 q^{\mathrm{hand}}_0(z)
 =\operatorname{clip}_{[q_{\min},q_{\max}]}
   (\bar q^{\mathrm{hand}}_0+z_0),\qquad
 v^{\mathrm{hand}}_0=0.
 \label{eq:joint-start}
\end{equation}
Here $\bar q^{\mathrm{hand}}_0$ is the reference hand configuration
expressed in the same pose and joint coordinates as the position-target
controls; $q_{\min}$ and $q_{\max}$ are componentwise bounds, and
$\operatorname{clip}$ enforces those bounds. Each first-window candidate
therefore starts from its own hand configuration, with zero hand
velocity and objects at their reference state. Subsequent windows start
from the robot and object states reached by execution. This couples
initial contact selection to its motion consequences while preserving
the simulated state across replans.

\subsection{Reusing Plans and Particles Across Windows}
\label{sec:mpc-reuse}
Consecutive windows overlap, so the previous solution provides useful
controls for much of the next window. We exploit this overlap through
three complementary mechanisms.

\textbf{Warm-started priors.}
We shift the previous window's selected plan by the executed prefix
and use it to center the next window's search. The Gaussian AR(1)
construction of Appendix~\ref{app:prior} is applied to deviations from
this shifted plan, so samples are temporally correlated corrections
to a promising continuation. The prior widens toward the window end,
allowing more exploration where the previous plan supplies less
information.

\textbf{Continuity at the window boundary.}
After the first window, we pin the first knot to the offset carried
over from the end of the executed prefix. Replanning therefore preserves
the command at the shared boundary instead of introducing a jump.
This constraint is enforced in both full searches and refinement
windows; only the remaining offsets are optimized. It leaves the first
window free to optimize its initial hand configuration through
Eq.~\eqref{eq:joint-start}.

\textbf{Alternating full search and local refinement.}
Full GRASP-TO searches alternate with cheaper refinement windows.
A full search uses the shifted prior and applies the tempering,
resampling, and surrogate-guided rejuvenation of
Algorithm~\ref{alg:grasp} to $J_t$. A refinement window instead shifts
the previous window's candidate plans and re-evaluates them from the
current state. These new evaluations are necessary because both the
starting state and the reference segment have changed.

The refinement then samples a first generation of correlated local
perturbations around promising shifted plans. A second generation
selects parents from the retained candidates and the first-generation
children, allowing improvements found in the first generation to guide
further search. Both generations are scored with the current window
objective. Refinement windows run no tempering stages: they reuse and
improve existing candidates between full population searches.

\subsection{Selecting and Executing a Plan}
\label{sec:mpc-selection}
Before execution, we shortlist up to 50 distinct candidate plans and
re-simulate them from the current state. We select the plan with the
lowest mean finite rollout cost across its repeated evaluations. This
additional comparison uses the same reference segment and objective
for every candidate and reduces dependence on a single rollout
evaluation.

We then execute only the committed prefix in the simulator and append
its states and controls to the retargeted trajectory. The next window
starts from the resulting state, and the selected plan and candidate
population supply the next warm start or refinement. Repeating this
cycle until the final reference frame produces a single dynamically
consistent trajectory while adapting future commands to the contacts actually
realized.
Algorithm~\ref{alg:grasp-mpc} summarizes this procedure.

\begin{algorithm}[t]
\caption{\grasp{}-MPC: closed-loop motion retargeting}
\label{alg:grasp-mpc}
\begin{algorithmic}[1]
\Require Initial state $x_0$, reference $\bar x_{0:T}$, window size $K$, commit count $C$
\State Set window index $j\leftarrow0$, knot index $k\leftarrow0$, library $\mathcal L\leftarrow\varnothing$
\While{$k<T/\Delta$}
  \State Set $K_j\leftarrow\min(K,T/\Delta-k)$ and $C_j\leftarrow\min(C,K_j)$
  \State Set $t\leftarrow k\Delta$ and $H_t\leftarrow K_j\Delta$; form $J_t$ from the current state using Eq.~\eqref{eq:mpc-cost}
  \If{$j=0$}
    \State For dexterous hands, let each candidate's $z_0$ set its initial hand state by Eq.~\eqref{eq:joint-start}
  \Else
    \State Shift the previous selected plan by $C_{j-1}$ knots and repeat its last knot to form the prior center
    \State Fix the first knot to the previous plan's offset at the shared boundary
  \EndIf
  \State Construct the AR(1) window prior, allowing broader exploration toward the tail
  \If{$j$ is even}
    \State Run Algorithm~\ref{alg:grasp} on $J_t$; retain its candidates and stage populations in $\mathcal L$
  \Else
    \State Shift $\mathcal L$ by $C_{j-1}$ knots, append an AR(1)-mean tail, and re-evaluate it under $J_t$
    \State Add two generations of correlated perturbations around promising candidates
  \EndIf
  \State Re-evaluate up to 50 distinct leading candidates with repeated rollouts
  \State Select $z^\star$ with the lowest mean finite rollout cost
  \State Execute its first $C_j$ knots; append the resulting states and controls
  \State Set the current state to the reached state; store $z^\star$ and set $k\leftarrow k+C_j$, $j\leftarrow j+1$
\EndWhile
\State \Return The executed state and control trajectory
\end{algorithmic}
\end{algorithm}

    \section{Experiment: Dexterous Hands}
\label{sec:exp-hands}
We evaluate \grasp{} on dexterous manipulation, testing whether a common
state-tracking objective can refine kinematically retargeted human
demonstrations into physically executable robot motions. The experiments
assess retargeting quality on large demonstration collections, transfer
across hand embodiments, and the value of the resulting trajectories for
policy learning and real-robot execution.

\textbf{Implementation.}
We use GRASP-MPC with MuJoCo \citep{todorov2012mujoco} and its Warp
backend\footnote{\url{https://github.com/google-deepmind/mujoco_warp}.}.
All dexterous-hand retargeting experiments share the tracking objective
in Eq.~\eqref{eq:tracking} and optimizer settings, with the particle counts
specified in each table. The objective
tracks robot and object states without a contact reward, leaving the
optimizer to discover contacts through simulated interactions under
gravity. We jointly optimize the initial hand configurations and the
first control window to refine the starting grasp.
Appendix~\ref{app:protocol} gives the implementation details,
task-selection protocol, and metric definitions.

\subsection{Bimanual Retargeting on OakInk2}
\textbf{Setup.}
We evaluate on \OverallEvaluated{} bimanual demonstrations from
OakInk2 \citep{zhan2024oakink2}, retargeted to two Sharpa hands.
The collection includes cutting, pouring, stirring, uncapping, inserting,
and opening, with a mean duration of \FullDurationMean{}\,s. These tasks
require coordinated contacts between both hands and the manipulated
objects while preserving the demonstrated object motion.
A trajectory succeeds when its mean object position
error is below 10\,cm and its mean rotation error is below 0.5\,rad.
Runs that terminate before completing the trajectory count as failures.
Table~\ref{tab:main} compares against kinematic dex-retargeting, SPIDER,
and Do-as-I-Do.

\begin{table}[htbp]
\centering
\small
\caption{Bimanual retargeting with Sharpa hands. \grasp{} is evaluated on
1352 OakInk2 demonstrations. All sampling-based methods use $N=1{,}024$ particles.
$E_{\mathrm{pos}}$ and $E_{\mathrm{rot}}$ denote mean object position and
rotation errors.}
\label{tab:main}
\setlength{\tabcolsep}{3pt}
\begin{tabular}{lccc}
\toprule
Method & Succ.\ (\%) $\uparrow$ & $E_{\mathrm{pos}}$ (m) $\downarrow$ & $E_{\mathrm{rot}}$ (rad) $\downarrow$ \\
\midrule
Dex-retargeting~\cite{qin2023anyteleop} & 28.6 & 0.08 & 0.62 \\
SPIDER~\cite{pan2025spider} mjwp & 71.4 & 0.04 & 0.57 \\
SPIDER~\cite{pan2025spider} mjwp\_act & 77.1 & 0.04 & 0.42 \\
Do-as-I-Do~\cite{paliwal2026do} (Sharpa) & 81.0 & 0.03 & 0.15 \\
\quad w/o transition reward & 79.0 & 0.03 & 0.14 \\
\quad annealed sampling only & 72.0 & 0.08 & 0.32 \\
\midrule
\textbf{GRASP} & \textbf{\MainSuccessRate{}} & \textbf{0.03} & \textbf{0.13} \\
\bottomrule
\end{tabular}
\end{table}

\textbf{Results.}
\grasp{} succeeds on \MainSuccess{}/\OverallEvaluated{} demonstrations
(\MainSuccessRate{}\%). Mean object position and rotation errors are
\MainPosMean{}\,cm and \MainRotMean{}\,rad, with medians of
\MainPosMedian{}\,cm and \MainRotMedian{}\,rad. These mean errors
are well below the success thresholds. Baseline success
rates range from 28.6\% for kinematic retargeting to 81.0\% for Do-as-I-Do
(Table~\ref{tab:main}). \grasp{} recovers
bimanual interactions across diverse tasks using a shared tracking
objective without task-specific contact guidance.
We study the effect of the particle count on tracking accuracy and
runtime in Appendix~\ref{app:particle-ablation}.

\subsection{Cross-Embodiment Retargeting}
\label{sec:cross-embodiment}
\textbf{Setup.}
We next test whether the same configuration transfers to hands with
different kinematics. We apply GRASP-MPC to OakInk2 demonstrations
using bimanual Wuji and XHand models, which have 20 and 12 finger joints
per hand, respectively, compared with 22 for Sharpa. Each hand also has
a 6-DoF base. We change the robot model and its kinematic reference while
keeping the tracking objective and optimizer settings fixed.
Table~\ref{tab:cross} reports results on the same 1,352-task collection
for each embodiment.

\begin{table}[htbp]
\centering
\small
\caption{Cross-embodiment retargeting on OakInk2 with one optimizer
configuration with $N=2{,}000$ particles for all three hands on 1,352 tasks
per hand. Execution failures count as
failures in the success rate.
Finger DoF are per hand; every hand also has a 6-DoF base.
$E_{\mathrm{pos}}$ and $E_{\mathrm{rot}}$ are mean / median object
position and rotation errors over completed trajectories with finite metrics.}
\label{tab:cross}
\setlength{\tabcolsep}{3pt}
\begin{tabular}{lccccc}
\toprule
Hand & Finger DoF & Tasks & Succ.\ (\%) $\uparrow$ & $E_{\mathrm{pos}}$ (cm) $\downarrow$ & $E_{\mathrm{rot}}$ (rad) $\downarrow$ \\
\midrule
Sharpa & 22 & 1352 & \OverallSuccessRate{} & \OverallPosMean{} / \OverallPosMedian{} & \OverallRotMean{} / \OverallRotMedian{} \\
Wuji & 20 & 1352 & 92.4 & 2.87 / 2.02 & 0.117 / 0.086 \\
XHand & 12 & 1352 & 95.0 & 2.87 / 2.08 & 0.112 / 0.088 \\
\bottomrule
\end{tabular}
\end{table}

\textbf{Results.}
Success rates exceed 92\% for all three hands: 92.4\%
for Wuji, 95.0\% for XHand, and
\OverallSuccessRate{}\% for Sharpa. Wuji and XHand both achieve a mean object
position error of 2.87\,cm, with mean rotation errors of
0.117 and 0.112\,rad, respectively. A single tracking formulation and
optimizer configuration transfers across embodiments.

\begin{figure}[htbp]
\centering
\includegraphics[width=\linewidth]{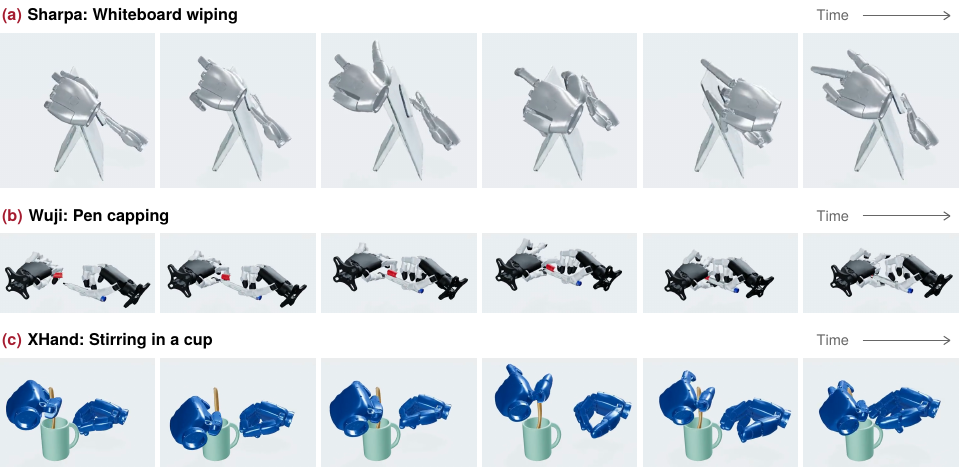}
\caption{Simulated dexterous manipulation across three hand embodiments.
Rows show (a) Sharpa wiping a whiteboard, (b) Wuji capping a pen, and
(c) XHand stirring in a cup. Each row shows six frames from one
retargeted OakInk2 trajectory, ordered from left to right, with a fixed crop.}
\label{fig:dex-sim}
\end{figure}

\textbf{Manipulation across embodiments.}
Figure~\ref{fig:dex-sim} shows three representative interactions:
Sharpa hands wipe a whiteboard, Wuji hands bring a pen and cap together,
and XHand hands stir in a cup. The sequences require sustained surface
contact, alignment of grasped parts, and coordinated tool--container
motion, respectively, illustrating the range of interactions supported
by the shared configuration.

\subsection{Retargeting HOT3D Demonstrations}
\textbf{Setup.}
To assess performance beyond OakInk2, we retarget \HotAllTasks{} HOT3D
trajectories \citep{banerjee2025hot3d} from the Aria and Quest3 recordings
to Sharpa hands. The collection includes right-hand, left-hand, and
bimanual interactions. We measure object tracking with ADD-AUC, which
integrates the fraction of frames satisfying an object-surface alignment
threshold as the threshold varies from 0 to 10\,cm. This metric captures
translation and rotation in a single score without fixing one threshold.
Table~\ref{tab:hot3d} includes the right-hand results reported by
SPIDER \citep{pan2025spider} alongside our breakdown by interaction type;
each row identifies its evaluation set.

\begin{table}[htbp]
\centering
\small
\caption{Object tracking on HOT3D. ADD-AUC integrates the fraction of
frames whose mean object-surface point error falls below a threshold
over thresholds of 0--10\,cm, normalized to a percentage. We use 512
area-uniform surface samples and average trajectories equally. \grasp{}
uses Sharpa hands on the evaluated Aria and Quest3 clips.}
\label{tab:hot3d}
\setlength{\tabcolsep}{6pt}
\begin{tabular}{llccc}
\toprule
 & & & & ADD-AUC (\%) $\uparrow$ \\
\cmidrule(lr){5-5}
Method & Split & \# Demos & \# Objects & Sharpa \\
\midrule
SPIDER \citep{pan2025spider} & Right hand & 2,718 & 33 & 40.74 \\
\midrule
\textbf{GRASP} & All & 5,400 & 33 & \textbf{81.54} \\
 & Right hand & 2,117 & 33 & 84.87 \\
 & Left hand & 1,213 & 33 & 85.44 \\
 & Bimanual & 2,070 & 33 & 75.83 \\
\bottomrule
\end{tabular}
\end{table}

\textbf{Results.}
\grasp{} achieves \HotAllAUC{}\% ADD-AUC over the full collection.
Right- and left-hand trajectories attain similar scores
(\HotRightAUC{}\% and \HotLeftAUC{}\%), while bimanual trajectories reach
\HotBimanualAUC{}\%, the lowest score among the three interaction types
in this collection. A single configuration thus supports thousands of
demonstrations from both recording platforms, including one- and
two-hand interactions. SPIDER reports 40.74\% for Sharpa on its right-hand evaluation set.

\subsection{Policy Learning from Retargeted Trajectories}
\label{sec:policy-learning}
\begin{figure}[htbp]
\centering
\includegraphics[width=\linewidth]{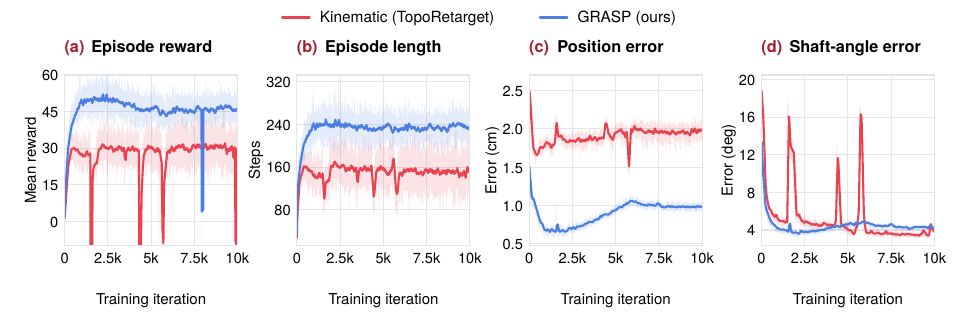}
\caption{PPO learning on nine pen-spinning clips using
kinematic references (TopoRetarget) or \grasp{}-refined references
(ours). Panels show (a) episode reward, (b) episode length, (c) object
position error, and (d) shaft-angle error. Bold curves are 100-iteration moving averages of the training logs;
faint curves are the unsmoothed values.}
\label{fig:rl-single-axis}
\end{figure}

\textbf{Setup.}
We assess the downstream value of reference refinement through pen spinning. On a group of nine clips, we train one PPO
policy to track the original kinematic references and one to track
their \grasp{}-optimized counterparts.
Both conditions use the same robot, reward configuration, domain
randomization, and training budget, varying only the reference.
Figure~\ref{fig:rl-single-axis} reports episode
reward, episode length, and object tracking errors during training.

\textbf{Results.}
Over the final 500 training iterations, the policy trained on \grasp{}
references achieves a mean episode reward of 45.9, compared with 21.7 for the kinematic references. Mean episode length
increases from 151 to 235 steps, and position error decreases from
1.98 to 0.99\,cm. Shaft-angle error increases slightly, from 3.9 to
4.3\,deg. The gains therefore lie in longer tracking episodes and lower
object position error, with a small tradeoff in shaft-angle accuracy.
The learning curves show these benefits over much of training,
supporting the use of physics-refined references for this task.

\subsection{Real-Robot Execution}
\begin{figure}[htbp]
\centering
\includegraphics[width=\linewidth]{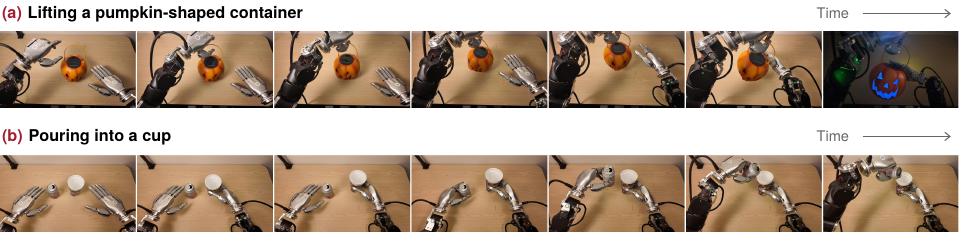}
\caption{Open-loop trajectory playback with real dexterous hands. (a) Lifting a
pumpkin-shaped container by its handle. (b) Pouring from a can into
a cup. Frames progress from left to right.}
\label{fig:realrobot}
\end{figure}

We execute \grasp{}-generated trajectories on physical dexterous hands
through open-loop playback. Figure~\ref{fig:realrobot} shows two
manipulation sequences: lifting a pumpkin-shaped container by its
handle and pouring from a can into a cup. The first requires supporting
an object through a narrow handle; the second requires coordinating
the container orientations during pouring. These demonstrations show
physical execution of the optimized motions for two distinct grasp
geometries and object interactions. Additional demonstrations are
available on the project website.

\section{Experiment: Humanoids}
\label{sec:exp-humanoid}
We next evaluate humanoid motion, which requires coordinating
whole-body movement, ground contact, and object manipulation. We measure
retargeting accuracy on the Unitree G1, illustrate interactions across
three humanoid embodiments, and assess the effect of reference refinement
on policy learning. Real-robot demonstrations examine crawling and box
manipulation.

\subsection{Whole-Body Motion Retargeting}
\textbf{Setup.}
We retarget motions from LAFAN1 \citep{harvey2020lafan},
AMASS \citep{mahmood2019amass}, and OMOMO \citep{li2023omomo} to the
Unitree G1 using GRASP-MPC. LAFAN1 and AMASS evaluate robot motion
tracking, while OMOMO also includes box manipulation. We report joint
and pelvis tracking errors on all three datasets and object pose errors
on OMOMO. Table~\ref{tab:humanoid} includes the GMR
\citep{araujo2025gmr} and SPIDER results reported by
\citet{pan2025spider}. Appendix~\ref{app:humanoid} describes the
evaluation and optimizer configuration.

\begin{table}[!htb]
\centering
\small
\caption{Whole-body tracking on the Unitree G1. $e_j$ denotes joint error;
$e_t$ and $e_r$ denote translation and orientation errors for the pelvis
and, on OMOMO, the object.}
\label{tab:humanoid}
\setlength{\tabcolsep}{6pt}
\begin{tabular}{lllccccc}
\toprule
 & & & \multicolumn{3}{c}{Robot} & \multicolumn{2}{c}{Object} \\
\cmidrule(lr){4-6}\cmidrule(lr){7-8}
Robot & Data & Method & $e_j\downarrow$ & $e_t\downarrow$ & $e_r\downarrow$ & $e_t\downarrow$ & $e_r\downarrow$ \\
\midrule
G1 & LAFAN1 & GMR~\cite{araujo2025gmr} & 1.08 & 2.01 & 2.40 & -- & -- \\
 &  & SPIDER~\cite{pan2025spider} & 0.58 & 0.11 & 0.07 & -- & -- \\
 &  & \textbf{GRASP} & 0.30 & 0.02 & 0.06 & -- & -- \\
\midrule
G1 & AMASS & GMR~\cite{araujo2025gmr} & 6.2 & 4.1 & 18.7 & -- & -- \\
 &  & SPIDER~\cite{pan2025spider} & 0.75 & 0.23 & 0.08 & -- & -- \\
 &  & \textbf{GRASP} & 0.59 & 0.05 & 0.12 & -- & -- \\
\midrule
G1 & OMOMO & GMR~\cite{araujo2025gmr} & 0.57 & 1.43 & 2.28 & 1.18 & 1.68 \\
 &  & SPIDER~\cite{pan2025spider} & 0.83 & 0.20 & 0.17 & 0.18 & 0.06 \\
 &  & \textbf{GRASP} & 0.21 & 0.06 & 0.06 & 0.03 & 0.13 \\
\bottomrule
\end{tabular}
\end{table}

\begin{figure}[H]
\centering
\includegraphics[width=0.88\linewidth]{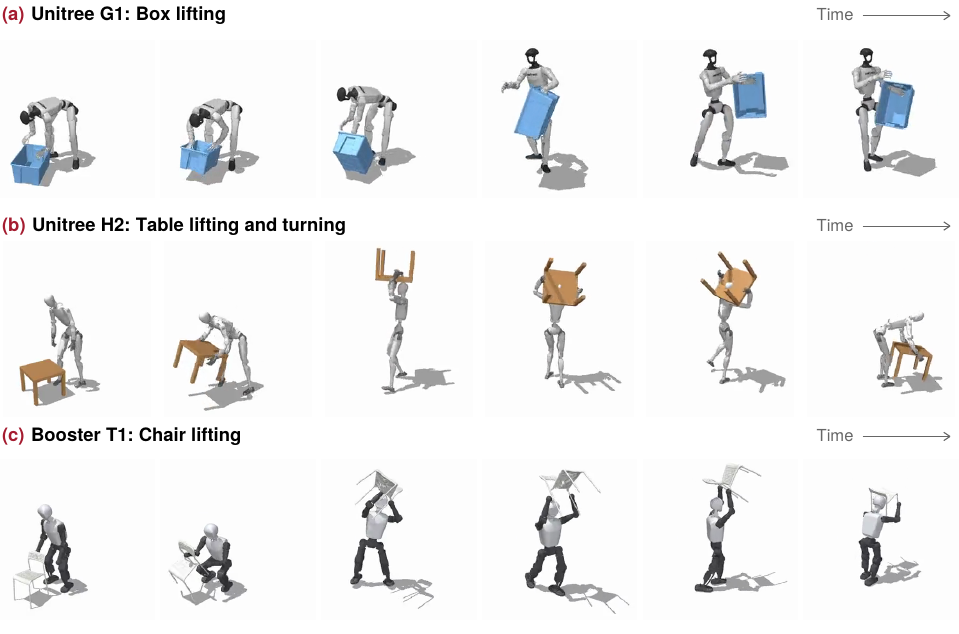}
\caption{Simulated humanoid--object interactions generated by \grasp{}.
Rows show (a) the Unitree G1 lifting a box, (b) the Unitree H2 lifting and turning
a table, and (c) the Booster T1 lifting a chair. Each row shows six
frames from one trajectory, ordered from left to right.}
\label{fig:humanoid-sim}
\end{figure}

\textbf{Results.}
Table~\ref{tab:humanoid} summarizes robot tracking on all three datasets
and joint robot--object tracking on OMOMO. For OMOMO, the recorded
\grasp{} errors are 0.21 for the joints, 0.06 for pelvis translation,
and 0.03 for object translation.

\textbf{Interactions across embodiments.}
Figure~\ref{fig:humanoid-sim} illustrates object manipulation with three
humanoid models. The Unitree G1 lifts a box from the floor, the Unitree H2
lifts and turns a table, and the Booster T1 raises a chair overhead.
These sequences require coordinated reaching, object support, and
changes in body posture. They illustrate the range of whole-body
interactions produced by the formulation beyond the G1 quantitative
evaluation.

\subsection{Learning Humanoid Policies from Refined References}
\label{sec:humanoid-rl}
\textbf{Setup.}
Following the downstream RL setup of DynaRetarget
\citep{dhedin2026dynaretarget}, we train PPO tracking policies for the
Unitree G1 using three reference motions from
OmniRetarget \citep{yang2025omniretarget}. For each motion, we compare
policies trained on the original kinematic trajectory with policies
trained on its \grasp{}-refined counterpart. Both conditions use the same
PPO configuration, reward weights, and domain randomization. We report
object position and orientation rewards, each measured relative to the
corresponding training reference, to assess how refinement affects
learning to track robot--object interactions.

\begin{figure}[H]
\centering
\includegraphics[width=0.85\linewidth]{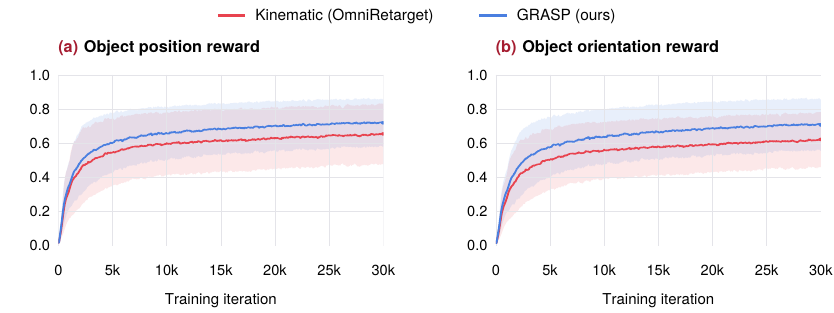}
\caption{Humanoid PPO learning with three reference motions from
OmniRetarget. (a) Object position and (b) orientation tracking
rewards are compared for the original kinematic trajectories (red) and
\grasp{}-refined trajectories (blue). Higher rewards indicate better
tracking of the corresponding reference.}
\label{fig:humanoid-rl}
\end{figure}

\textbf{Results.}
Figure~\ref{fig:humanoid-rl} shows that \grasp{} references yield higher
mean object position and orientation rewards through most of training.
The curves separate early and remain separated at the end of the
30,000-iteration budget: the refined motions provide more learnable
tracking references. Together with the pen-spinning results, these
experiments show that physics-based refinement improves downstream policy
learning in both dexterous-hand and humanoid tasks.

\subsection{Real-Robot Execution}
\begin{figure}[H]
\centering
\includegraphics[width=0.95\linewidth]{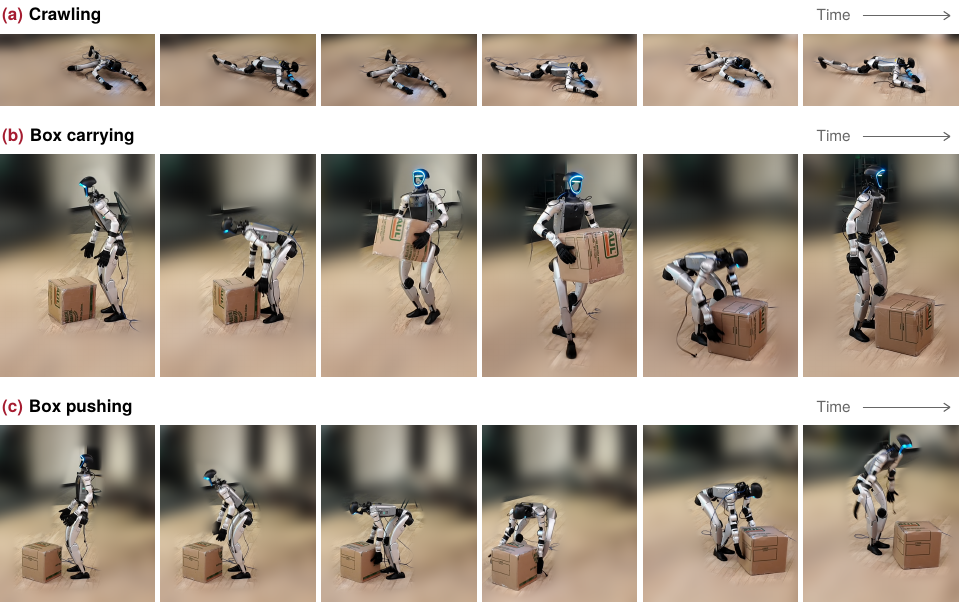}
\caption{Closed-loop execution of learned policies on real humanoids.
Rows show (a) crawling,
(b) lifting, carrying, and placing a box, and (c) pushing a box along
the floor. Each row shows six frames from one execution, ordered from
left to right. Backgrounds are blurred for clarity, preserving the robot,
manipulated object, and immediate contact regions.}
\label{fig:humanoid-real}
\end{figure}

We deploy tracking policies trained on \grasp{}-refined references to
control real humanoids in closed loop; deployment follows OmniRetarget
\citep{yang2025omniretarget}. Figure~\ref{fig:humanoid-real}
shows crawling, box carrying, and box pushing. Crawling requires
sustained limb--floor contact. During box carrying, the robot bends down,
lifts the box, walks while supporting it, and returns it to the floor.
During box pushing, the robot reaches down and moves the box along the
floor before releasing it. These demonstrations cover ground contact,
object support during locomotion, and object displacement through hand
contact.

\FloatBarrier
\section{Conclusion}
\label{sec:conclusion}
We presented \grasp{}, a method that combines sampling-based exploration
and gradient-guided refinement for physics-based motion retargeting
without differentiable simulator dynamics. GRASP-TO recovers guidance
from local quadratic models of archived rollouts, with a Metropolis
correction that preserves the original target for fixed proposal
parameters. Our analysis relates the population fit to a Gaussian-smoothed
cost, while GRASP-MPC extends the solver to long demonstrations by reusing
plans across overlapping control windows.

Experiments show that a shared state-tracking objective can recover
contacts across dexterous-hand and humanoid demonstrations without
task-specific contact rewards. The refined references improve downstream
PPO tracking, and hardware demonstrations include dexterous manipulation,
humanoid crawling, and box manipulation. These results support rollout
reuse as a practical source of optimization guidance and training
references for robot learning. Future work will study the reliability of
guidance from finite, regularized archives and the robustness of the
resulting motions and policies to errors in object geometry and dynamics.

\setlength{\bibsep}{2pt plus 1pt minus 1pt}
\bibliographystyle{plainnat}
\bibliography{grasp_references}

\begin{thebibliography}{55}
\providecommand{\natexlab}[1]{#1}
\providecommand{\url}[1]{\texttt{#1}}
\expandafter\ifx\csname urlstyle\endcsname\relax
  \providecommand{\doi}[1]{doi: #1}\else
  \providecommand{\doi}{doi: \begingroup \urlstyle{rm}\Url}\fi

\bibitem[Araujo et~al.(2025)Araujo, Ze, Xu, Wu, and Liu]{araujo2025gmr}
Joao~Pedro Araujo, Yanjie Ze, Pei Xu, Jiajun Wu, and C.~Karen Liu.
\newblock Retargeting matters: General motion retargeting for humanoid motion
  tracking.
\newblock \emph{arXiv preprint arXiv:2510.02252}, 2025.
\newblock URL \url{https://arxiv.org/abs/2510.02252}.

\bibitem[Banerjee et~al.(2025)Banerjee, Shkodrani, Moulon, Hampali, Han, Zhang,
  Zhang, Fountain, Miller, Basol, Newcombe, Wang, Engel, and
  Hodan]{banerjee2025hot3d}
Prithviraj Banerjee, Sindi Shkodrani, Pierre Moulon, Shreyas Hampali, Shangchen
  Han, Fan Zhang, Linguang Zhang, Jade Fountain, Edward Miller, Selen Basol,
  Richard Newcombe, Robert Wang, Jakob~Julian Engel, and Tomas Hodan.
\newblock {HOT3D}: Hand and object tracking in {3D} from egocentric multi-view
  videos.
\newblock In \emph{Proceedings of the IEEE/CVF Conference on Computer Vision
  and Pattern Recognition (CVPR)}, 2025.
\newblock URL \url{https://arxiv.org/abs/2411.19167}.

\bibitem[Berzuini and Gilks(2001)]{berzuini2001resample}
Carlo Berzuini and Walter~R. Gilks.
\newblock {RESAMPLE-MOVE} filtering with cross-model jumps.
\newblock In Arnaud Doucet, Nando de~Freitas, and Neil Gordon, editors,
  \emph{Sequential Monte Carlo Methods in Practice}, pages 117--138. Springer,
  2001.
\newblock \doi{10.1007/978-1-4757-3437-9_6}.

\bibitem[Beskos et~al.(2016)Beskos, Jasra, Kantas, and
  Thiery]{beskos2016adaptive}
Alexandros Beskos, Ajay Jasra, Nikolas Kantas, and Alexandre Thiery.
\newblock On the convergence of adaptive sequential {Monte Carlo} methods.
\newblock \emph{The Annals of Applied Probability}, 26\penalty0 (2):\penalty0
  1111--1146, 2016.
\newblock \doi{10.1214/15-AAP1113}.
\newblock URL \url{https://arxiv.org/abs/1306.6462}.

\bibitem[Bharadhwaj et~al.(2020)Bharadhwaj, Xie, and
  Shkurti]{bharadhwaj2020mpc}
Homanga Bharadhwaj, Kevin Xie, and Florian Shkurti.
\newblock Model-predictive control via cross-entropy and gradient-based
  optimization.
\newblock In \emph{Proceedings of the 2nd Conference on Learning for Dynamics
  and Control}, volume 120 of \emph{Proceedings of Machine Learning Research},
  pages 277--286, 2020.
\newblock URL \url{https://proceedings.mlr.press/v120/bharadhwaj20a.html}.

\bibitem[Cheng et~al.(2024)Cheng, Ji, Chen, Yang, Yang, and
  Wang]{cheng2024exbody}
Xuxin Cheng, Yandong Ji, Junming Chen, Ruihan Yang, Ge~Yang, and Xiaolong Wang.
\newblock Expressive whole-body control for humanoid robots.
\newblock \emph{arXiv preprint arXiv:2402.16796}, 2024.
\newblock URL \url{https://arxiv.org/abs/2402.16796}.

\bibitem[Conn et~al.(2009)Conn, Scheinberg, and Vicente]{conn2009dfo}
Andrew~R. Conn, Katya Scheinberg, and Luis~N. Vicente.
\newblock \emph{Introduction to Derivative-Free Optimization}.
\newblock SIAM, 2009.
\newblock \doi{10.1137/1.9780898718768}.

\bibitem[Conrad et~al.(2016)Conrad, Marzouk, Pillai, and
  Smith]{conrad2016accelerating}
Patrick~R. Conrad, Youssef~M. Marzouk, Natesh~S. Pillai, and Aaron Smith.
\newblock Accelerating asymptotically exact {MCMC} for computationally
  intensive models via local approximations.
\newblock \emph{Journal of the American Statistical Association}, 111\penalty0
  (516):\penalty0 1591--1607, 2016.
\newblock \doi{10.1080/01621459.2015.1096787}.
\newblock URL \url{https://arxiv.org/abs/1402.1694}.

\bibitem[Del~Moral et~al.(2006)Del~Moral, Doucet, and Jasra]{delmoral2006smc}
Pierre Del~Moral, Arnaud Doucet, and Ajay Jasra.
\newblock Sequential {Monte Carlo} samplers.
\newblock \emph{Journal of the Royal Statistical Society: Series B},
  68\penalty0 (3):\penalty0 411--436, 2006.
\newblock URL
  \url{https://www.stats.ox.ac.uk/~doucet/delmoral_doucet_jasra_sequentialmontecarlosamplersJRSSB.pdf}.

\bibitem[Dellaert(2003)]{dellaert2003sample}
Frank Dellaert.
\newblock A sample of {Monte Carlo} methods in robotics and vision.
\newblock In \emph{Proceedings of the ISM International Symposium on the
  Science of Modeling}, 2003.
\newblock URL \url{https://dellaert.github.io/files/Dellaert03ism.pdf}.

\bibitem[Dellaert et~al.(1999)Dellaert, Fox, Burgard, and
  Thrun]{dellaert1999monte}
Frank Dellaert, Dieter Fox, Wolfram Burgard, and Sebastian Thrun.
\newblock {Monte Carlo} localization for mobile robots.
\newblock In \emph{IEEE International Conference on Robotics and Automation},
  volume~2, pages 1322--1328, 1999.
\newblock \doi{10.1109/ROBOT.1999.772544}.
\newblock URL
  \url{https://sites.cc.gatech.edu/aimosaic/robot-lab/online-publications/dellaert_frank_1999_2.pdf}.

\bibitem[Deutscher et~al.(2000)Deutscher, Blake, and
  Reid]{deutscher2000articulated}
Jonathan Deutscher, Andrew Blake, and Ian Reid.
\newblock Articulated body motion capture by annealed particle filtering.
\newblock In \emph{IEEE Conference on Computer Vision and Pattern Recognition},
  volume~2, pages 126--133, 2000.
\newblock \doi{10.1109/CVPR.2000.854758}.
\newblock URL
  \url{https://www.robots.ox.ac.uk/~vdg/abstracts/cvpr2000-jdeutsch.html}.

\bibitem[Dh{\'e}din et~al.(2026)Dh{\'e}din, Taouil, Omar, Yu, Tao, Dai, and
  Khadiv]{dhedin2026dynaretarget}
Victor Dh{\'e}din, Ilyass Taouil, Shafeef Omar, Dian Yu, Kun Tao, Angela Dai,
  and Majid Khadiv.
\newblock {DynaRetarget}: Dynamically-feasible retargeting using sampling-based
  trajectory optimization.
\newblock \emph{arXiv preprint arXiv:2602.06827}, 2026.
\newblock URL \url{https://arxiv.org/abs/2602.06827}.

\bibitem[Esteban et~al.(2026)Esteban, Siu, Mach, Li, Kurtz, Burdick, and
  Ames]{esteban2026shooting}
Sergio~A. Esteban, Jason H.~K. Siu, Derrick Mach, Junheng Li, Vince Kurtz,
  Joel~W. Burdick, and Aaron~D. Ames.
\newblock Shooting for contact: Contact-implicit multiple shooting for dynamic
  motion retargeting.
\newblock \emph{arXiv preprint arXiv:2608.03116}, 2026.
\newblock URL \url{https://arxiv.org/abs/2608.03116}.

\bibitem[Gilks and Berzuini(2001)]{gilks2001moving}
Walter~R. Gilks and Carlo Berzuini.
\newblock Following a moving target---{Monte Carlo} inference for dynamic
  {Bayesian} models.
\newblock \emph{Journal of the Royal Statistical Society: Series B},
  63\penalty0 (1):\penalty0 127--146, 2001.
\newblock \doi{10.1111/1467-9868.00280}.

\bibitem[Graesdal et~al.(2024)Graesdal, Chia, Marcucci, Morozov, Amice,
  Parrilo, and Tedrake]{graesdal2024tight}
Bernhard~Paus Graesdal, Shao Yuan~Chew Chia, Tobia Marcucci, Savva Morozov,
  Alexandre Amice, Pablo~A. Parrilo, and Russ Tedrake.
\newblock Towards tight convex relaxations for contact-rich manipulation.
\newblock In \emph{Robotics: Science and Systems}, 2024.
\newblock \doi{10.15607/RSS.2024.XX.132}.
\newblock URL \url{https://arxiv.org/abs/2402.10312}.

\bibitem[Groudiev et~al.(2025)Groudiev, Kang, and Yang]{groudiev2025cuadmm}
Antoine Groudiev, Shucheng Kang, and Heng Yang.
\newblock {cuADMM}: {GPU}-accelerated first-order optimization for large-scale
  multi-block semidefinite programs.
\newblock In \emph{RSS Workshop on Fast Motion Planning and Control in the Era
  of Parallelism}, 2025.
\newblock URL
  \url{https://agroudiev.github.io/assets/pdf/8_cuADMM_GPU_Accelerated_First.pdf}.

\bibitem[Harvey et~al.(2020)Harvey, Yurick, Nowrouzezahrai, and
  Pal]{harvey2020lafan}
F{\'e}lix~G. Harvey, Mike Yurick, Derek Nowrouzezahrai, and Christopher Pal.
\newblock Robust motion in-betweening.
\newblock \emph{ACM Transactions on Graphics (SIGGRAPH)}, 2020.
\newblock URL \url{https://arxiv.org/abs/2102.04942}.

\bibitem[Hastings(1970)]{hastings1970monte}
W.~K. Hastings.
\newblock Monte carlo sampling methods using markov chains and their
  applications.
\newblock \emph{Biometrika}, 57\penalty0 (1):\penalty0 97--109, 1970.
\newblock \doi{10.1093/biomet/57.1.97}.

\bibitem[He et~al.(2024)He, Luo, Xiao, Zhang, Kitani, Liu, and Shi]{he2024h2o}
Tairan He, Zhengyi Luo, Wenli Xiao, Chong Zhang, Kris Kitani, Changliu Liu, and
  Guanya Shi.
\newblock Learning human-to-humanoid real-time whole-body teleoperation.
\newblock \emph{arXiv preprint arXiv:2403.04436}, 2024.
\newblock URL \url{https://arxiv.org/abs/2403.04436}.

\bibitem[He et~al.(2025)He, Gao, Xiao, Zhang, Wang, Wang, Luo, He, Sobanbab,
  Pan, Yi, Qu, Kitani, Hodgins, Fan, Zhu, Liu, and Shi]{he2025asap}
Tairan He, Jiawei Gao, Wenli Xiao, Yuanhang Zhang, Zi~Wang, Jiashun Wang,
  Zhengyi Luo, Guanqi He, Nikhil Sobanbab, Chaoyi Pan, Zeji Yi, Guannan Qu,
  Kris Kitani, Jessica Hodgins, Linxi Fan, Yuke Zhu, Changliu Liu, and Guanya
  Shi.
\newblock {ASAP}: Aligning simulation and real-world physics for learning agile
  humanoid whole-body skills.
\newblock In \emph{Robotics: Science and Systems (RSS)}, 2025.
\newblock URL \url{https://arxiv.org/abs/2502.01143}.

\bibitem[Honda et~al.(2023)Honda, Akai, Suzuki, Aoki, Hosogaya, Okuda, and
  Suzuki]{honda2023svgmppi}
Kohei Honda, Naoki Akai, Kosuke Suzuki, Mizuho Aoki, Hirotaka Hosogaya,
  Hiroyuki Okuda, and Tatsuya Suzuki.
\newblock Stein variational guided model predictive path integral control:
  Proposal and experiments with fast maneuvering vehicles.
\newblock \emph{arXiv preprint arXiv:2309.11040}, 2023.
\newblock URL \url{https://arxiv.org/abs/2309.11040}.

\bibitem[Kalakrishnan et~al.(2011)Kalakrishnan, Chitta, Theodorou, Pastor, and
  Schaal]{kalakrishnan2011stomp}
Mrinal Kalakrishnan, Sachin Chitta, Evangelos Theodorou, Peter Pastor, and
  Stefan Schaal.
\newblock {STOMP}: Stochastic trajectory optimization for motion planning.
\newblock In \emph{IEEE International Conference on Robotics and Automation},
  2011.
\newblock URL \url{https://is.mpg.de/publications/kalakrishnan_raiic_2011}.

\bibitem[Kang et~al.(2025)Kang, Liu, and Yang]{kang2025global}
Shucheng Kang, Guorui Liu, and Heng Yang.
\newblock Global contact-rich planning with sparsity-rich semidefinite
  relaxations.
\newblock In \emph{Robotics: Science and Systems}, 2025.
\newblock \doi{10.15607/RSS.2025.XXI.046}.
\newblock URL \url{https://arxiv.org/abs/2502.02829}.

\bibitem[Kang et~al.(2026)Kang, Xu, Sarva, Liang, and Yang]{kang2026fast}
Shucheng Kang, Xiaoyang Xu, Jay Sarva, Ling Liang, and Heng Yang.
\newblock Fast and certifiable trajectory optimization.
\newblock In \emph{Algorithmic Foundations of Robotics XVI, Volume 1},
  volume~37 of \emph{Springer Proceedings in Advanced Robotics}, pages 43--65.
  Springer, 2026.
\newblock \doi{10.1007/978-3-032-09967-9_3}.
\newblock URL
  \url{https://link.springer.com/chapter/10.1007/978-3-032-09967-9_3}.

\bibitem[Khan et~al.(2004)Khan, Balch, and Dellaert]{khan2004mcmc}
Zia Khan, Tucker Balch, and Frank Dellaert.
\newblock An {MCMC}-based particle filter for tracking multiple interacting
  targets.
\newblock In \emph{European Conference on Computer Vision}, pages 279--290,
  2004.
\newblock URL \url{https://dellaert.github.io/files/Khan04eccv.pdf}.

\bibitem[Li et~al.(2023)Li, Wu, and Liu]{li2023omomo}
Jiaman Li, Jiajun Wu, and C.~Karen Liu.
\newblock Object motion guided human motion synthesis.
\newblock \emph{ACM Transactions on Graphics (SIGGRAPH Asia)}, 2023.
\newblock URL \url{https://arxiv.org/abs/2309.16237}.

\bibitem[Li et~al.(2026)Li, Gong, and Chalvatzaki]{li2026impact}
Jiayun Li, Dejian Gong, and Georgia Chalvatzaki.
\newblock {IMPACT}: An implicit active-set augmented lagrangian for fast
  contact-implicit trajectory optimization.
\newblock In \emph{Robotics: Science and Systems}, 2026.
\newblock \doi{10.15607/RSS.2026.XXII.163}.

\bibitem[Li et~al.(2025)Li, Han, Kang, Ma, and Yang]{li2025crisp}
Yulin Li, Haoyu Han, Shucheng Kang, Jun Ma, and Heng Yang.
\newblock On the surprising robustness of sequential convex optimization for
  contact-implicit motion planning.
\newblock In \emph{Robotics: Science and Systems}, 2025.
\newblock \doi{10.15607/RSS.2025.XXI.047}.

\bibitem[Mahmood et~al.(2019)Mahmood, Ghorbani, Troje, Pons-Moll, and
  Black]{mahmood2019amass}
Naureen Mahmood, Nima Ghorbani, Nikolaus~F. Troje, Gerard Pons-Moll, and
  Michael~J. Black.
\newblock {AMASS}: Archive of motion capture as surface shapes.
\newblock In \emph{Proceedings of the IEEE/CVF International Conference on
  Computer Vision (ICCV)}, 2019.
\newblock URL \url{https://arxiv.org/abs/1904.03278}.

\bibitem[Mordatch et~al.(2012)Mordatch, Popovic, and Todorov]{mordatch2012hand}
Igor Mordatch, Zoran Popovic, and Emanuel Todorov.
\newblock Contact-invariant optimization for hand manipulation.
\newblock In \emph{Eurographics/ACM SIGGRAPH Symposium on Computer Animation},
  pages 137--144, 2012.
\newblock \doi{10.2312/SCA/SCA12/137-144}.

\bibitem[Nesterov and Spokoiny(2017)]{nesterov2017random}
Yurii Nesterov and Vladimir Spokoiny.
\newblock Random gradient-free minimization of convex functions.
\newblock \emph{Foundations of Computational Mathematics}, 17\penalty0
  (2):\penalty0 527--566, 2017.
\newblock \doi{10.1007/s10208-015-9296-2}.

\bibitem[Oh et~al.(2005)Oh, Rehg, Balch, and Dellaert]{oh2005datadriven}
Sang~Min Oh, James~M. Rehg, Tucker Balch, and Frank Dellaert.
\newblock Data-driven {MCMC} for learning and inference in switching linear
  dynamic systems.
\newblock In \emph{Proceedings of the Twentieth National Conference on
  Artificial Intelligence}, pages 944--949, 2005.
\newblock URL \url{https://cdn.aaai.org/AAAI/2005/AAAI05-149.pdf}.

\bibitem[Paliwal et~al.(2026)Paliwal, Etukuru, Liang, Abbeel, Shafiullah, and
  Malik]{paliwal2026do}
Bhawna Paliwal, Haritheja Etukuru, William Liang, Pieter Abbeel, Nur
  Muhammad~Mahi Shafiullah, and Jitendra Malik.
\newblock Do as i do: Dexterous manipulation data from everyday human videos.
\newblock \emph{arXiv preprint arXiv:2606.19333}, 2026.
\newblock URL \url{https://arxiv.org/abs/2606.19333}.

\bibitem[Pan et~al.(2025)Pan, Wang, Qi, Urain, Liu, Bharadhwaj, Sharma, Wu,
  Shi, Malik, and Hogan]{pan2025spider}
Chaoyi Pan, Changhao Wang, Haozhi Qi, Julen Urain, Zixi Liu, Homanga
  Bharadhwaj, Akash Sharma, Tingfan Wu, Guanya Shi, Jitendra Malik, and
  Francois Hogan.
\newblock {SPIDER}: Scalable physics-informed dexterous retargeting.
\newblock \emph{arXiv preprint arXiv:2511.09484}, 2025.
\newblock URL \url{https://arxiv.org/abs/2511.09484}.

\bibitem[Park et~al.(2026)Park, Kim, Lee, and Choi]{park2026nfg}
Kisang Park, Chan~Woo Kim, Kyungjae Lee, and Sungjoon Choi.
\newblock Natural functional gradients for smooth trajectory optimization.
\newblock In \emph{Robotics: Science and Systems}, 2026.
\newblock \doi{10.15607/RSS.2026.XXII.162}.

\bibitem[Pinneri et~al.(2021)Pinneri, Sawant, Blaes, Achterhold, Stueckler,
  Rolinek, and Martius]{pinneri2021icem}
Cristina Pinneri, Shambhuraj Sawant, Sebastian Blaes, Jan Achterhold, Joerg
  Stueckler, Michal Rolinek, and Georg Martius.
\newblock Sample-efficient cross-entropy method for real-time planning.
\newblock In \emph{Proceedings of the 2020 Conference on Robot Learning},
  volume 155 of \emph{Proceedings of Machine Learning Research}, pages
  1049--1065, 2021.
\newblock URL \url{https://proceedings.mlr.press/v155/pinneri21a.html}.

\bibitem[Posa et~al.(2014)Posa, Cantu, and Tedrake]{posa2014direct}
Michael Posa, Cecilia Cantu, and Russ Tedrake.
\newblock A direct method for trajectory optimization of rigid bodies through
  contact.
\newblock \emph{The International Journal of Robotics Research}, 33\penalty0
  (1):\penalty0 69--81, 2014.
\newblock \doi{10.1177/0278364913506757}.

\bibitem[Qin et~al.(2021)Qin, Wu, Liu, Jiang, Yang, Fu, and Wang]{qin2022dexmv}
Yuzhe Qin, Yueh-Hua Wu, Shaowei Liu, Hanwen Jiang, Ruihan Yang, Yang Fu, and
  Xiaolong Wang.
\newblock {DexMV}: Imitation learning for dexterous manipulation from human
  videos.
\newblock \emph{arXiv preprint arXiv:2108.05877}, 2021.
\newblock URL \url{https://arxiv.org/abs/2108.05877}.

\bibitem[Qin et~al.(2023)Qin, Yang, Huang, Van~Wyk, Su, Wang, Chao, and
  Fox]{qin2023anyteleop}
Yuzhe Qin, Wei Yang, Binghao Huang, Karl Van~Wyk, Hao Su, Xiaolong Wang, Yu-Wei
  Chao, and Dieter Fox.
\newblock {AnyTeleop}: A general vision-based dexterous robot arm-hand
  teleoperation system.
\newblock In \emph{Robotics: Science and Systems}, 2023.

\bibitem[Rakita et~al.(2025)Rakita, Liang, and Wang]{rakita2025wasp}
Daniel Rakita, Chen Liang, and Qian Wang.
\newblock Coherence-based approximate derivatives via web of affine spaces
  optimization.
\newblock In \emph{Robotics: Science and Systems}, 2025.
\newblock \doi{10.15607/RSS.2025.XXI.024}.

\bibitem[Ranganathan and Dellaert(2005)]{ranganathan2005datadriven}
Ananth Ranganathan and Frank Dellaert.
\newblock Data driven {MCMC} for appearance-based topological mapping.
\newblock In \emph{Robotics: Science and Systems}, 2005.
\newblock \doi{10.15607/RSS.2005.I.028}.
\newblock URL \url{https://www.roboticsproceedings.org/rss01/p28.html}.

\bibitem[Rasmussen(2003)]{rasmussen2003gaussian}
Carl~Edward Rasmussen.
\newblock Gaussian processes to speed up hybrid {Monte Carlo} for expensive
  {Bayesian} integrals.
\newblock In \emph{Bayesian Statistics 7}, pages 651--659. Oxford University
  Press, 2003.
\newblock URL \url{https://mlg.eng.cam.ac.uk/pub/pdf/Ras03.pdf}.

\bibitem[Roberts and Tweedie(1996)]{roberts1996langevin}
Gareth~O. Roberts and Richard~L. Tweedie.
\newblock Exponential convergence of langevin distributions and their discrete
  approximations.
\newblock \emph{Bernoulli}, 2\penalty0 (4):\penalty0 341--363, 1996.
\newblock \doi{10.2307/3318418}.

\bibitem[Rubinstein(1999)]{rubinstein1999cem}
Reuven Rubinstein.
\newblock The cross-entropy method for combinatorial and continuous
  optimization.
\newblock \emph{Methodology and Computing in Applied Probability}, 1\penalty0
  (2):\penalty0 127--190, 1999.
\newblock \doi{10.1023/A:1010091220143}.

\bibitem[Suh et~al.(2022{\natexlab{a}})Suh, Pang, and Tedrake]{suh2022bundled}
H.~J.~Terry Suh, Tao Pang, and Russ Tedrake.
\newblock Bundled gradients through contact via randomized smoothing.
\newblock \emph{IEEE Robotics and Automation Letters}, 2022{\natexlab{a}}.
\newblock URL \url{https://arxiv.org/abs/2109.05143}.

\bibitem[Suh et~al.(2026)Suh, Pang, Zhao, and Tedrake]{suh2026ctr}
H.~J.~Terry Suh, Tao Pang, Tong Zhao, and Russ Tedrake.
\newblock Dexterous contact-rich manipulation via the contact trust region.
\newblock \emph{The International Journal of Robotics Research}, 45\penalty0
  (9):\penalty0 1418--1454, 2026.
\newblock \doi{10.1177/02783649251398875}.

\bibitem[Suh et~al.(2022{\natexlab{b}})Suh, Simchowitz, Zhang, and
  Tedrake]{suh2022policy}
Hyung~Ju Suh, Max Simchowitz, Kaiqing Zhang, and Russ Tedrake.
\newblock Do differentiable simulators give better policy gradients?
\newblock In \emph{Proceedings of the 39th International Conference on Machine
  Learning}, volume 162 of \emph{Proceedings of Machine Learning Research},
  pages 20668--20696, 2022{\natexlab{b}}.
\newblock URL \url{https://proceedings.mlr.press/v162/suh22b.html}.

\bibitem[Sun et~al.(2026)Sun, Jordana, Fornasier, Etesami, and
  Khadiv]{sun2026cbo}
Xudong Sun, Armand Jordana, Massimo Fornasier, Jalal Etesami, and Majid Khadiv.
\newblock Consensus-based optimization ({CBO}): Towards global optimality in
  robotics.
\newblock In \emph{Robotics: Science and Systems}, 2026.
\newblock \doi{10.15607/RSS.2026.XXII.161}.

\bibitem[Todorov et~al.(2012)Todorov, Erez, and Tassa]{todorov2012mujoco}
Emanuel Todorov, Tom Erez, and Yuval Tassa.
\newblock {MuJoCo}: A physics engine for model-based control.
\newblock In \emph{IEEE/RSJ International Conference on Intelligent Robots and
  Systems}, pages 5026--5033, 2012.
\newblock \doi{10.1109/IROS.2012.6386109}.

\bibitem[Williams et~al.(2017)Williams, Wagener, Goldfain, Drews, Rehg, Boots,
  and Theodorou]{williams2017mppi}
Grady Williams, Nolan Wagener, Brian Goldfain, Paul Drews, James~M. Rehg, Byron
  Boots, and Evangelos~A. Theodorou.
\newblock Information theoretic {MPC} for model-based reinforcement learning.
\newblock In \emph{IEEE International Conference on Robotics and Automation},
  2017.
\newblock \doi{10.1109/ICRA.2017.7989202}.
\newblock URL
  \url{https://homes.cs.washington.edu/~bboots/files/InformationTheoreticMPC.pdf}.

\bibitem[Yang(2026)]{yang2026tempered}
Heng Yang.
\newblock Tempered sequential monte carlo for trajectory and policy
  optimization with differentiable dynamics.
\newblock In \emph{Robotics: Science and Systems}, 2026.
\newblock \doi{10.15607/RSS.2026.XXII.109}.
\newblock URL \url{https://arxiv.org/abs/2604.21456}.

\bibitem[Yang et~al.(2025{\natexlab{a}})Yang, Huang, Wu, Kanazawa, Abbeel,
  Sferrazza, Liu, Duan, and Shi]{yang2025omniretarget}
Lujie Yang, Xiaoyu Huang, Zhen Wu, Angjoo Kanazawa, Pieter Abbeel, Carmelo
  Sferrazza, C.~Karen Liu, Rocky Duan, and Guanya Shi.
\newblock {OmniRetarget}: Interaction-preserving data generation for humanoid
  whole-body loco-manipulation and scene interaction.
\newblock \emph{arXiv preprint arXiv:2509.26633}, 2025{\natexlab{a}}.
\newblock URL \url{https://arxiv.org/abs/2509.26633}.

\bibitem[Yang et~al.(2025{\natexlab{b}})Yang, Suh, Zhao, Graesdal, Kelestemur,
  Wang, Pang, and Tedrake]{yang2025physicsgen}
Lujie Yang, H.~J.~Terry Suh, Tong Zhao, Bernhard~Paus Graesdal, Tarik
  Kelestemur, Jiuguang Wang, Tao Pang, and Russ Tedrake.
\newblock Physics-driven data generation for contact-rich manipulation via
  trajectory optimization.
\newblock In \emph{Robotics: Science and Systems}, 2025{\natexlab{b}}.
\newblock \doi{10.15607/RSS.2025.XXI.053}.

\bibitem[Zhan et~al.(2024)Zhan, Yang, Zhao, Mao, Xu, Lin, Li, and
  Lu]{zhan2024oakink2}
Xinyu Zhan, Lixin Yang, Yifei Zhao, Kangrui Mao, Hanlin Xu, Zenan Lin, Kailin
  Li, and Cewu Lu.
\newblock {OAKINK2}: A dataset of bimanual hands-object manipulation in complex
  task completion.
\newblock In \emph{Proceedings of the IEEE/CVF Conference on Computer Vision
  and Pattern Recognition}, 2024.
\newblock URL \url{https://arxiv.org/abs/2403.19417}.

\end{thebibliography}

\clearpage
\beginappendix
\begingroup
\small
\section{AR(1) Prior}
\label{app:prior}
The Gaussian autoregressive prior $p_0(z)$ correlates successive knot
offsets within each control channel while keeping channels independent.
For channel $c\in\{1,\ldots,m\}$ in Eq.~\eqref{eq:knots}, we sample
\begin{equation}
 z_{0,c}\sim\mathcal N\!\Bigl(0,\tfrac{s_{0,c}^2}{1-\rho^2}\Bigr),\qquad
 z_{k,c}=\rho\,z_{k-1,c}+s_{k,c}\,\xi_{k,c},\quad
 \xi_{k,c}\sim\mathcal N(0,1),
 \label{eq:ar1}
\end{equation}
for $k=1,\ldots,K-1$, with independent standard-normal perturbations
$\xi_{k,c}$. Here $\rho\in[0,1)$ controls temporal correlation, and
$s_{k,c}=\sigma\kappa_{k,c}$ is the exploration scale, with base scale
$\sigma>0$ and channel- and knot-dependent factors $\kappa_{k,c}>0$.
The corresponding negative log density is
\begin{equation}
 -\log p_0(z)=\frac12\sum_{c=1}^{m}\Bigl[
   \frac{(1-\rho^2)\,z_{0,c}^2}{s_{0,c}^2}
   +\sum_{k=1}^{K-1}\frac{(z_{k,c}-\rho z_{k-1,c})^2}{s_{k,c}^2}\Bigr]
   +\mathrm{const}.
 \label{eq:ar1-density}
\end{equation}
With knot-independent scales $s_{k,c}=s_c$, the initialization in
Eq.~\eqref{eq:ar1} makes the process stationary, with variance
$s_c^2/(1-\rho^2)$ and correlation $\rho^{|k-k'|}$ between knots $k$ and
$k'$. In the dexterous-hand experiments, we use $\rho=0.9$,
$\sigma=0.0044$, and a factor of 10 on finger channels. For MPC windows
with $K>1$ knots, we additionally multiply the scales by
$4^{k/(K-1)}$ to increase exploration toward the window end. As described
in Section~\ref{sec:mpc-reuse}, the MPC prior is centered on the shifted
previous plan, and the boundary knot is fixed after the first window.

\textbf{Interpretation.}
Each knot retains a fraction $\rho$ of the preceding offset and adds a
perturbation with standard deviation $s_{k,c}$. The stationary prior
therefore favors sustained control corrections: with $\rho=0.9$ and
0.2\,s knot spacing, offsets one second apart have correlation
$0.9^5\approx0.59$. Before window-dependent widening or boundary
conditioning, the nominal stationary standard deviations are about
1\,cm for hand-base positions, 0.01\,rad for base rotations, and 0.1\,rad
for finger joints. The larger finger scale allows more variation in
grasp configuration, while the base remains closer to the reference.
Widening the MPC prior provides additional freedom where the shifted
plan supplies less information. The prior covariance also preconditions
the Langevin proposals, giving their perturbations the same temporal
structure.

\section{Gradient Recovery and Metropolis Correction}
\label{app:theory}
\subsection{Derivatives of a Smoothed Objective}
We first justify replacing full rollout costs with the reduced response
in Eq.~\eqref{eq:reduced-response}. Let $Z\sim\nu_z$,
$Y=B^\top D^{-1}(Z-\mu)$, and $F(Y)=\mathbb E[J(Z)\mid Y]$.
Assume $\mathbb E[J(Z)^2]<\infty$. For any square-integrable predictor
$h(Y)$, conditioning on $Y$ gives
\begin{equation}
 \mathbb E[(J(Z)-h(Y))^2]
 =\mathbb E[\operatorname{Var}(J(Z)\mid Y)]
  +\mathbb E[(F(Y)-h(Y))^2].
\end{equation}
The conditional-variance term is independent of $h$. Consequently,
fitting a quadratic to $J(Z)$ or to $F(Y)$ yields the same population
minimizer.

We prove Proposition~\ref{prop:mollified} under its stated Gaussian
fitting-measure and polynomial-growth assumptions.
\begin{proof}
Express the quadratic in the centered Gaussian polynomial basis as
$a^\circ+b^\top Y+\tfrac12\operatorname{tr}[A(YY^\top-\Sigma)]$,
where $a^\circ=a+\tfrac12\operatorname{tr}(A\Sigma)$.
The constant, linear, and centered quadratic blocks are orthogonal in
$L^2(\nu)$. Their normal equations give
\begin{align}
 b_*&=\Sigma^{-1}\mathbb E[YF(Y)],\\
 A_*&=\Sigma^{-1}\mathbb E[(YY^\top-\Sigma)F(Y)]\Sigma^{-1}.
\end{align}
The remaining coefficient is
$a_*=\mathbb E[F(Y)]-\tfrac12\operatorname{tr}(A_*\Sigma)$.
Since $\Sigma\succ0$, these normal equations determine a unique
quadratic. Differentiating the Gaussian convolution kernel yields the
same expressions for $\nabla\overline F(0)$ and
$\nabla^2\overline F(0)$, establishing Eq.~\eqref{eq:mollified}.
Polynomial growth ensures integrability and permits differentiation
under the integral; no derivative of $F$ is required.
\end{proof}

\subsection{Fixed-Kernel Metropolis Correction}
The surrogate determines where the Markov kernel proposes to move,
while the acceptance test uses the original rollout objective. The
following statement formalizes why fitting error does not change the
invariant distribution of a fixed kernel.
\begin{propositionbox}
\begin{proposition}[Fixed-kernel correction]
\label{prop:fixed-kernel}
Let $J$ be a deterministic rollout objective with normalizable target
$\pi_\lambda$ in Eq.~\eqref{eq:target}. Fix $\lambda$, the surrogate
(including $\mu,D,B$), the preconditioner $M\succ0$, and the step size
$\epsilon>0$. The Markov kernel defined by the proposal in
Eq.~\eqref{eq:mala}, surrogate drift in Eq.~\eqref{eq:guided}, and
acceptance rule in Eq.~\eqref{eq:mh} is reversible with respect to
$\pi_\lambda$, regardless of the surrogate fitting error.
\end{proposition}
\end{propositionbox}
\begin{proof}
For distinct states $z$ and $z'$, the accepted-transition density satisfies
\begin{align*}
\pi_\lambda(z)Q_\lambda(z'\mid z)\alpha(z,z')
&=\min\{\pi_\lambda(z)Q_\lambda(z'\mid z),
          \pi_\lambda(z')Q_\lambda(z\mid z')\}\\
&=\pi_\lambda(z')Q_\lambda(z\mid z')\alpha(z',z).
\end{align*}
This is detailed balance. Rejection leaves the state unchanged, so the
diagonal part of the transition also satisfies detailed balance.
The full kernel is therefore reversible and leaves $\pi_\lambda$
invariant.
\end{proof}

\section{Dexterous-Hand Evaluation Details}
\label{app:protocol}
\textbf{Task collection.}
We retarget OakInk2 \citep{zhan2024oakink2} demonstrations to two Sharpa
hands in MuJoCo Warp. The main collection contains \OverallEvaluated{}
clips, with mean and median durations of \FullDurationMean{}\,s and
\FullDurationMedian{}\,s. Seven clips were used during initial
development; at $N=2{,}000$, excluding them gives \NoDevSuccess{} successes out of
\NoDevEvaluated{} clips (\NoDevSuccessRate{}\%). All 1,352 tasks are included
in the main results, with execution failures counted as failures.

\textbf{Controls and objective.}
The bimanual Sharpa model has $m=56$ control channels: two 6-DoF hand
bases and 44 finger joints. In Eq.~\eqref{eq:tracking}, the object
position and rotation weights are 1 and 0.3; the hand-base position and
rotation weights are 0.1 and 0.03; and the finger-joint weight is 0.003.
We set $w_v=0.0001$ and $w_{\rm term}=1$, with no contact reward.
Objects evolve under nominal gravity and contact throughout optimization.
The initial hand configuration is optimized jointly with the first
control window using Eq.~\eqref{eq:joint-start}.

\textbf{Optimizer settings.}
Table~\ref{tab:hand-settings} lists the optimizer configuration used for
dexterous-hand evaluation.

\begin{table}[htbp]
\centering\small
\caption{GRASP-MPC settings for dexterous-hand retargeting.}
\label{tab:hand-settings}
\setlength{\tabcolsep}{4pt}
\renewcommand{\arraystretch}{1.08}
\begin{tabularx}{\columnwidth}{@{}lX@{}}
\toprule
Setting & Value \\
\midrule
Timing & 0.005\,s simulation step; 0.2\,s knot spacing \\
MPC window & Up to 15 knots; commit count $C=2$ \\
Particles & $N=1{,}024$ in Table~\ref{tab:main}; $N=2{,}000$ for all hands in
Table~\ref{tab:cross}; ESS target $0.9N$ \\
SMC moves & 5 Markov moves per stage; at most 64 tempering stages \\
Temperature & $\beta=T_{\mathrm{clip}}/\tau_0$;
$\tau_0=0.006$\,s \\
Prior & Appendix~\ref{app:prior} \\
Surrogate & $r=2$ temporal basis components per channel; at most 60,000
archived evaluations; cross-validated guidance shrinkage \\
Reuse and selection & Sections~\ref{sec:mpc-reuse}
and~\ref{sec:mpc-selection} \\
Seed and hardware & Seed 0; NVIDIA A100 80\,GB GPU \\
\bottomrule
\end{tabularx}
\end{table}

Let $T_{\mathrm{clip}}$ denote the clip duration and $\Delta t$ the
simulation timestep. We normalize each window objective by the number
of steps in the full clip: $J=(\Delta t/T_{\mathrm{clip}})\sum_t\ell_t$,
where the sum includes the stage and weighted terminal tracking terms.
Setting $\beta=T_{\mathrm{clip}}/\tau_0$ with
$\tau_0=0.006\,\mathrm{s}$ gives
\begin{equation}
 \beta J=\frac{\Delta t}{\tau_0}\sum_t \ell_t
\end{equation}
so the coefficient multiplying the window's accumulated loss does not
depend on the full clip duration.

\textbf{Metrics.}
For object $o$ at time $t$, let $p_{o,t}\in\mathbb R^3$ and
$R_{o,t}\in\mathrm{SO}(3)$ denote the simulated position and orientation,
with bars marking their reference values. The tracking errors are
\begin{align}
 e^p_{o,t}&=\|p_{o,t}-\bar p_{o,t}\|_2,\\
 e^R_{o,t}&=\|\operatorname{Log}(\bar R_{o,t}^{\top}R_{o,t})\|_2.
\end{align}
Here $\operatorname{Log}$ maps a relative rotation to its axis--angle
vector, so $e^R_{o,t}$ is measured in radians. Following the common
object-tracking metric implementation, if one reference object's mean
translational displacement is below 1\,mm, we use the other object's
errors; otherwise, we average both objects. Averaging over the
demonstrated motion segment gives the task-level errors
$E_{\mathrm{pos}}$ and $E_{\mathrm{rot}}$. Success requires
$E_{\mathrm{pos}}<0.1$\,m and $E_{\mathrm{rot}}<0.5$\,rad; runs that
terminate before completing the segment count as failures.

\textbf{HOT3D.}
We use the same Sharpa models, objective, and GRASP-MPC settings for
HOT3D. The evaluation contains 2,960 Aria and 2,440 Quest3 trajectories,
including right-hand, left-hand, and bimanual interactions, scored at
50\,Hz. As defined in Table~\ref{tab:hot3d}, ADD-AUC integrates
object-surface alignment over thresholds from 0 to 10\,cm and weights
trajectories equally.

\textbf{Pen-spinning PPO.}
We adapt wuji-mjlab v2026.10.22 (commit \texttt{b2b98d7}) to a right Sharpa
hand with 22 finger joints, keeping its original gains, armature, collision
shapes, and contact pairs. The wrist is fixed and references are expressed
in the palm frame, so training tracks finger and object motion without the
moving-wrist dynamics of the source trajectories. Wuji-specific torque and
overcurrent penalties, fingertip-force multipliers, effort-limit
randomization, and pen resizing are removed; PPO, tracking rewards, adaptive
sampling, the remaining domain randomization, and the disturbance curriculum
are retained. Kinematic references are the Sharpa IK
trajectories, and \grasp{} references are the executed trajectories sampled
from 200 to 50\,Hz. Each condition trains one policy shared across the nine
clips, with 4,096 parallel environments, 10,000 PPO iterations, and seed 0.

\textbf{Real-hand playback.}
Optimized trajectories are replayed open loop at 50\,Hz on OpenArm arms with
Sharpa Wave hands.

\section{Humanoid Evaluation Details}
\label{app:humanoid}
\textbf{Evaluation.}
We evaluate robot tracking on LAFAN1, AMASS, and OMOMO, with additional
object tracking on OMOMO. For clip $c$ with $n_c$ frames, let
$\theta_{c,t}\in\mathbb R^{29}$ denote the robot joint angles,
$p_{c,t}$ the pelvis position, and $r_{c,t}$ its unit quaternion; starred
quantities denote the corresponding reference states. We compute
\begin{align}
e_{j,c} &= \frac{1}{n_c}\sum_t
  \|\theta_{c,t}-\theta^*_{c,t}\|_2, &
e_{t,c} &= \frac{1}{n_c}\sum_t
  \|p_{c,t}-p^*_{c,t}\|_2,\\
e_{r,c} &= \frac{1}{n_c}\sum_t
  2\arccos\!\left(\min\{1,|\langle r_{c,t},r^*_{c,t}\rangle|\}\right).
\end{align}
The joint error is the per-frame $\ell_2$ norm over all 29 joints. Joint
and rotation errors are reported in radians, and translation errors in
meters. The absolute quaternion inner product makes $e_{r,c}$ invariant
to the signs of the two quaternion representations. Object translation
and rotation errors on OMOMO use the same definitions. We compare each
executed frame with its corresponding 60\,Hz reference frame, average
over frames within each clip, and then average the clip-level errors so
that every clip has equal weight.

\textbf{Optimizer settings.}
The Unitree G1 has $m=29$ actuated joints. Table~\ref{tab:humanoid-settings}
lists the complete optimizer configuration used for evaluation.

\begin{table}[htbp]
\centering\small
\caption{GRASP-MPC settings for humanoid tracking. The LAFAN1/AMASS weight
tuple is (pelvis translation, pelvis rotation, joint angles, contact); the
OMOMO tuple is (pelvis translation, pelvis rotation, joint angles, object
translation, object rotation).}
\label{tab:humanoid-settings}
\setlength{\tabcolsep}{4pt}
\renewcommand{\arraystretch}{1.08}
\begin{tabularx}{\columnwidth}{@{}lX@{}}
\toprule
Setting & Value \\
\midrule
Simulator & MJWarp; direct PD control; no domain randomization \\
Timing & $1/60$\,s simulation step; 0.1\,s knot spacing \\
MPC window & 8 knots; commit count $C=1$; first knot pinned \\
Particles & $N=1{,}000$; ESS target $0.9N$ \\
SMC moves & Surrogate MALA; full-prior preconditioner; 5 Markov moves per
stage; no fixed stage cap \\
Surrogate & Rank 2; ridge $10^{-3}$; window parameter 3 \\
Tracking weights & LAFAN1/AMASS: $(3,1,1,1)$; OMOMO:
$(1,1,1,1,0.3)$; velocity: 0; intermediate terminal cost: 1 \\
Temperature convention & $\beta=1/\tau$ \\
LAFAN1 & $\tau=1.82\times10^{-4}$ ($\beta=5.49\times10^3$);
$(\rho,\sigma)=(0.9,0.0142)$; initial step size 0.0026; arm scale 1 \\
AMASS & $\tau=4.55\times10^{-5}$ ($\beta=2.20\times10^4$);
$(\rho,\sigma)=(0,0.065)$; initial step size 0.02; arm scale 0.5 \\
OMOMO & $\tau=1.82\times10^{-4}$ ($\beta=5.49\times10^3$);
$(\rho,\sigma)=(0,0.065)$; initial step size 0.02; arm scale 0.5 \\
Step adaptation & Target 0.05; rate 3; minimum scale 0.1 \\
Warm start & Shift previous plan; widen the prior exploration scale toward
the tail by a factor of 4 \\
Library and selection & Reuse every 2 windows with two-generation local
improvement; re-evaluate the best 50 distinct candidates with 20 rollouts
each and execute the lowest-mean-cost candidate \\
\bottomrule
\end{tabularx}
\end{table}

All runs use random seed 0, eight CPU cores, and one NVIDIA A100-SXM4-80GB GPU.

\section{Trajectory Optimization Benchmarks}
\label{app:benchmarks}
\label{sec:benchmarks}
These benchmarks isolate GRASP-TO's behavior under smooth dynamics and
discrete contact transitions, complementing the retargeting experiments
in Sections~\ref{sec:exp-hands} and~\ref{sec:exp-humanoid}.

\textbf{Setup.}
We compare GRASP-TO with gradient-based TSMC \citep{yang2026tempered},
random-walk Metropolis TSMC (RWM-TSMC), and Parallel-MPPI, a parallel
baseline that runs an independent MPPI instance per particle
\citep{williams2017mppi}. Acrobot swing-up has smooth dynamics, allowing
the gradient-based comparator to differentiate through the simulator.
The planar pusher-slider selects a discrete contact edge at each step;
for this task, the comparator uses a straight-through Gumbel--Softmax
relaxation to obtain gradients. All reported costs are evaluated with
the original simulator. Table~\ref{tab:benchmark-settings} lists the problem
dimensions and optimization budgets.

\begin{table}[htbp]
\centering\small
\caption{Settings for the trajectory optimization benchmarks.}
\label{tab:benchmark-settings}
\setlength{\tabcolsep}{4pt}
\renewcommand{\arraystretch}{1.08}
\begin{tabularx}{\columnwidth}{@{}lX@{}}
\toprule
Setting & Value \\
\midrule
Problem sizes & Both tasks: $H=200$, $N=100$; control dimension 1
(acrobot) or 4 (pusher-slider), giving 200 or 800 variables \\
Gradient TSMC & 3 NUTS transitions per stage; at most 8 tree doublings
and 200 adaptive tempering stages \\
GRASP-TO & 99 proposals per particle and stage; at most 200 adaptive
tempering stages \\
RWM-TSMC & 99 proposals per particle and stage; at most 200 adaptive
tempering stages \\
Parallel-MPPI & 100 independent particles; 64 rollouts per update;
15 pusher-slider updates; acrobot updates matched to the per-particle
rollout budget of rollout-based TSMC \\
Hardware & NVIDIA A100 80\,GB GPU \\
Runtime boundary & From optimizer entry to return; includes each method's
first-call JAX compilation; excludes common initialization, initial
cost-function compilation, final re-scoring, and plotting \\
\bottomrule
\end{tabularx}
\end{table}

\begin{table}[htbp]
\centering\small
\caption{Smooth and nonsmooth trajectory optimization benchmarks.
Cost reports the mean $\pm$ standard deviation, across seeds, of the
lowest-cost particle in each final population. We use seven seeds for
acrobot and eleven for the pusher-slider. A pusher-slider run is solved
when its lowest-cost final box pose is within 10\,cm and 0.5\,rad of the
goal;
time is measured in wall-clock seconds.}
\label{tab:optimization}
\setlength{\tabcolsep}{9pt}
\begin{tabular}{lccccc}
\toprule
 & \multicolumn{2}{c}{Acrobot} & \multicolumn{3}{c}{Pusher-slider} \\
\cmidrule(lr){2-3}\cmidrule(lr){4-6}
Method & Cost $\downarrow$ & Time (s) & Cost $\downarrow$ & Success & Time (s) \\
\midrule
Gradient-based TSMC & $2534\pm119$ & 2069 & $63.7\pm4.4$ & 11/11 & 3075 \\
\textbf{GRASP-TO} & $\mathbf{2504\pm98}$ & 46 & $\mathbf{54.3\pm2.6}$ & 11/11 & 354 \\
RWM-TSMC & $2787\pm210$ & 39 & $117.3\pm9.7$ & 1/11 & 71 \\
Parallel-MPPI & $3117\pm150$ & 2 & $188.6\pm15.2$ & 0/11 & 2 \\
\bottomrule
\end{tabular}
\end{table}

\begin{figure}[H]
\centering
\includegraphics[width=\linewidth]{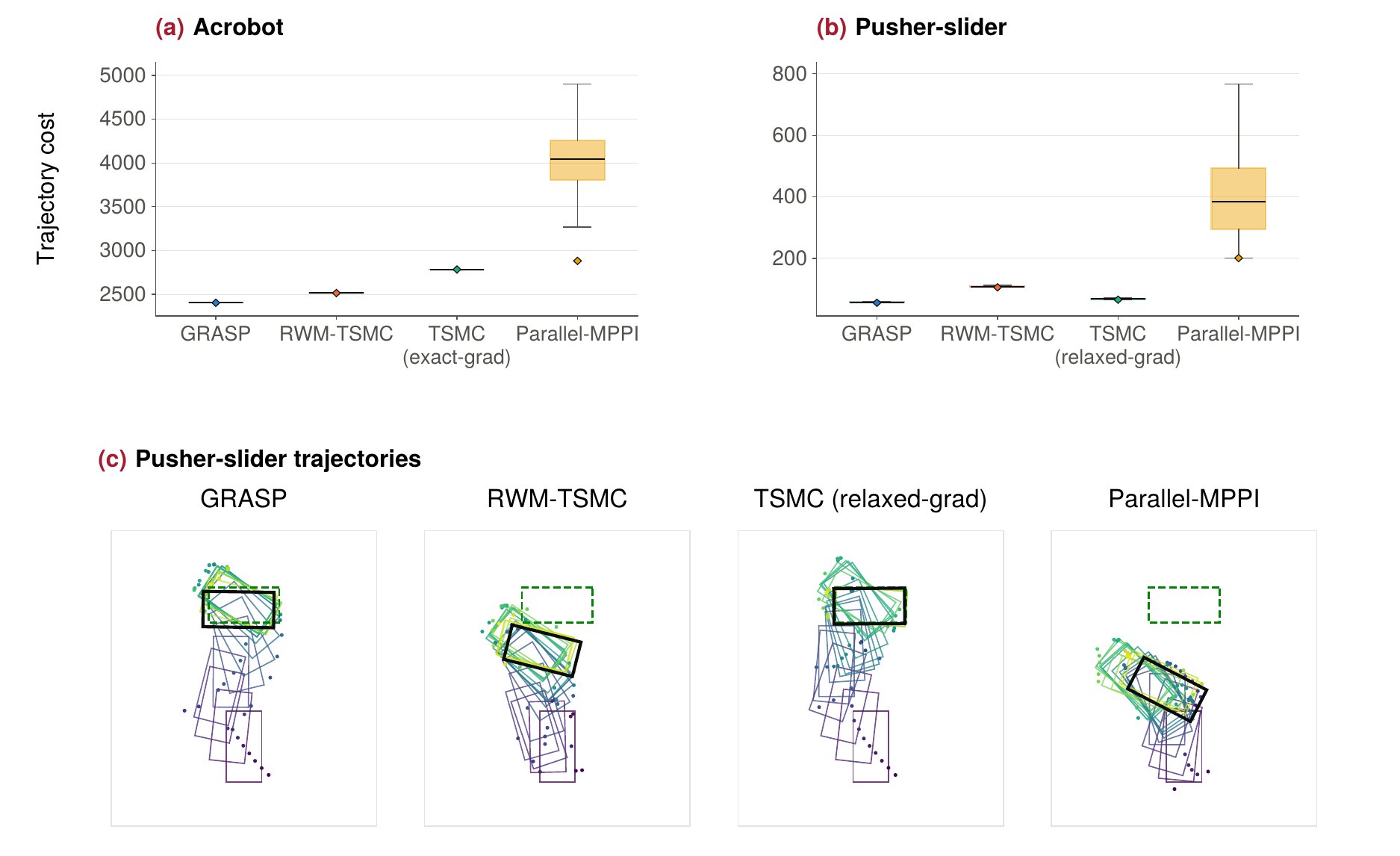}
\caption{Cost distributions of the final particle populations for representative runs on
(a) acrobot and (b) pusher-slider; diamonds mark the best particles.
(c) Pusher-slider trajectories, colored by time, with dashed goal poses.
The gradient-based TSMC comparator uses simulator derivatives for acrobot
and a contact relaxation for the pusher-slider.}
\label{fig:optimization}
\end{figure}

\textbf{Results.}
On acrobot, GRASP-TO reaches a cost of $2504\pm98$, comparable to
$2534\pm119$ for gradient-based TSMC, with a reported runtime of 46\,s
versus 2069\,s (Table~\ref{tab:optimization}). On the pusher-slider,
GRASP-TO reaches $54.3\pm2.6$, compared with $63.7\pm4.4$ for the
gradient-based comparator. Both solve all eleven runs. RWM-TSMC and
Parallel-MPPI run faster in these configurations but attain higher costs;
RWM-TSMC solves one of the pusher-slider runs and Parallel-MPPI none.
Figure~\ref{fig:optimization} shows representative particle populations
and trajectories. Together, these results demonstrate that guidance
recovered from rollout costs can support both local refinement under
smooth dynamics and search across discrete contact transitions.

\FloatBarrier
\clearpage
\section{Particle Count Ablation}
\label{app:particle-ablation}
We compare $N\in\{512,1024,2000\}$ for Sharpa bimanual retargeting on
OakInk2, using one NVIDIA A100-SXM4-80GB GPU per run. Increasing the particle count reduces both mean and median tracking errors, while increasing median runtime.
Note that since the errors in Table~\ref{tab:particle-ablation} and
Figure~\ref{fig:particle-ablation} are computed on the 1,305 tasks with
finite metrics for all three particle counts, the $N=1{,}024$ and
$N=2{,}000$ errors differ slightly from those in Tables~\ref{tab:main}
and~\ref{tab:cross}, which use all tasks with finite metrics for each
particle count. Success rates are computed over all 1,352 tasks.

\begin{table}[htbp]
\centering\small
\caption{Particle count ablation on Sharpa. Position errors $e_{\mathrm{pos}}$
are in centimeters and rotation errors $e_{\mathrm{rot}}$ in radians.
Wall time measures the MPC loop from entry to return, excluding
initialization and final evaluation.}
\label{tab:particle-ablation}
\setlength{\tabcolsep}{7pt}
\begin{tabular}{rcccccc}
\toprule
$N$ & \multicolumn{2}{c}{$e_{\mathrm{pos}}$ (cm)}
    & \multicolumn{2}{c}{$e_{\mathrm{rot}}$ (rad)}
    & Wall time (min) & Success (\%) \\
\cmidrule(lr){2-3}\cmidrule(lr){4-5}
 & Mean & Median & Mean & Median & Median & \\
\midrule
512  & 3.274 & 2.504 & 0.1346 & 0.1068 & 30.98 & 93.12 \\
1024 & 3.164 & 2.301 & 0.1263 & 0.0982 & 39.43 & 92.75 \\
2000 & 2.848 & 2.162 & 0.1165 & 0.0905 & 57.49 & 94.16 \\
\bottomrule
\end{tabular}
\end{table}

\begin{figure}[htbp]
\centering
\includegraphics[width=\linewidth]{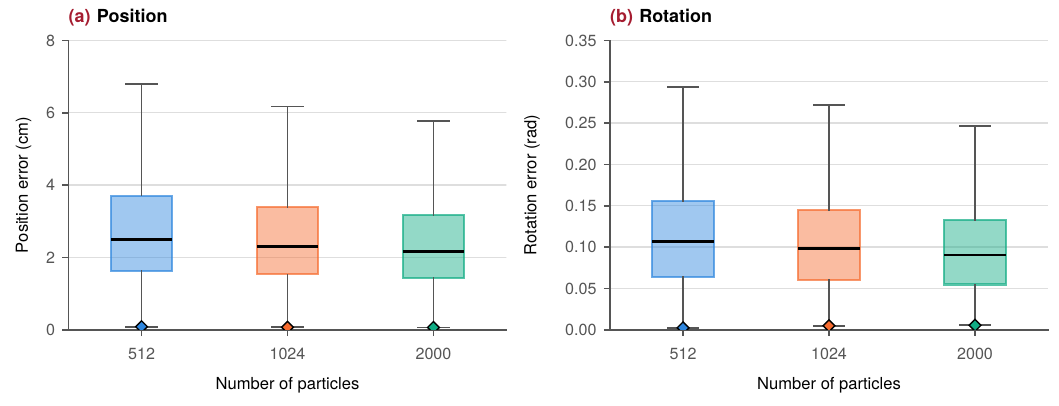}
\caption{Sharpa tracking errors for three particle counts:
(a) position and (b) rotation. Boxes span the interquartile range,
black lines mark medians, whiskers extend to observations within
1.5 interquartile ranges, and diamonds mark minimum task errors.}
\label{fig:particle-ablation}
\end{figure}

\FloatBarrier

\endgroup
\end{document}